\documentclass{bmvc2k}
\usepackage{graphicx}
\usepackage{booktabs}
\usepackage[accsupp]{axessibility}  
\usepackage{colortbl}
\usepackage{pifont}
\usepackage{multirow}
\usepackage{sansmath}
\usepackage{makecell}
\usepackage{paralist}
\usepackage{amsmath}
\usepackage{tikz}
\usepackage{inconsolata}
\usepackage{subcaption}
\usepackage{float}

\definecolor{color1bg}{HTML}{4285f4}
\definecolor{color2bg}{HTML}{ea4335}
\newcommand{\CC}{\cellcolor{LightCyan}}

\usepackage{hyperref}

\usepackage{dirtytalk}

\definecolor{Gray}{gray}{0.90}
\definecolor{white}{rgb}{1.0, 1.0, 1.0}

\definecolor{LightCyan}{RGB}{247, 223, 231}
\newcolumntype{a}{>{\columncolor{LightCyan}}c}
\definecolor{Gray}{gray}{0.90}

\def\thumos{THUMOS14\xspace}
\def\anet{ActivityNet\mbox{-1.3}\xspace}
\def\aformer{ActionFormer\xspace}
\def\ovformer{OVFormer\xspace}
\newcommand{\RNum}[1]{\uppercase\expandafter{\romannumeral #1\relax}}
\title{Open-Vocabulary Temporal Action Localization using Multimodal Guidance}

\addauthor{Akshita Gupta}{agupta22@uoguelph.ca}{1,3}
\addauthor{Aditya Arora}{adityac8@yorku.ca}{2,3}
\addauthor{Sanath Narayan}{sanath.narayan@tii.ae}{4}
\addauthor{Salman Khan}{salman.khan@mbzuai.ac.ae}{5}
\addauthor{Fahad Shahbaz Khan}{fahad.khan@mbzuai.ac.ae}{5}
\addauthor{Graham W.~Taylor}{gwtaylor@uoguelph.ca}{1,3}

\addinstitution{
 University of Guelph, \\
 Guelph, Ontario \\
}
\addinstitution{
 York University, \\
 Toronto, Ontario \\
}
\addinstitution{
 Vector Institute,\\
 Toronto, Ontario \\
}
\addinstitution{
 Technology Innovation Institute,\\
 Abu Dhabi, UAE \\
}
\addinstitution{
 Mohamed Bin Zayed University of Artificial Intelligence,\\
 Abu Dhabi, UAE \\
}

\runninghead{GUPTA ET.~AL}{OVTAL using Multimodal Guidance}

\begin{document}

\maketitle

\begin{abstract}

Open-Vocabulary Temporal Action Localization (OVTAL) enables a model to recognize any desired action category in videos without the need to explicitly curate training data for all categories. However, this flexibility poses significant challenges, as the model must recognize not only the action categories seen during training but also novel categories specified at inference. Unlike standard temporal action localization, where training and test categories are predetermined, OVTAL requires understanding contextual cues that reveal the semantics of novel categories. To address these challenges, we introduce \ovformer, a novel open-vocabulary framework extending \aformer with three key contributions. First, we employ task-specific prompts as input to a large language model to obtain rich class-specific descriptions for action categories. Second, we introduce a cross-attention mechanism to learn the alignment between class representations and frame-level video features, facilitating the multimodal guided features. Third, we propose a two-stage training strategy which includes training with a larger vocabulary dataset and finetuning to downstream data to generalize to novel categories. \ovformer extends existing TAL methods to open-vocabulary settings. Comprehensive evaluations on the \thumos and \anet benchmarks demonstrate the effectiveness of our method. Code and pretrained models will be publicly released.


\end{abstract}
\section{Introduction}
\label{sec:intro}
\vspace{-0.2cm}

Temporal action localization (TAL) aims to localize and classify every action instance in a long untrimmed video. This task is crucial for tasks such as video understanding, surveillance and summarizing videos. In recent years, numerous methods have emerged to address TAL \cite{chao2018rethinking, lin2021learning, liu2020progressive, zhao2021video}, achieving significant performance at localizing and recognizing a fixed set of action categories. However, most works are restricted to a closed-set setting.
To localize novel action categories unseen during training, these approaches require training the model on the combined set of base and novel categories using additional annotated instances from the novel classes under consideration. With the increasing volume of videos, annotating every action instance in videos is impractical.
In this work, we relax the restriction of localizing closed-set action classes in the TAL setting and propose an Open-Vocabulary TAL (OVTAL) approach, called OVFormer. Our OVFormer strives to localize both base actions defined during training as well as novel action classes during inference. 

Predicting novel classes during inference poses a significantly greater challenge compared to standard TAL or its closely related problems such as the open-set \cite{bao2022opental,chen2023cascade}, zero-shot \cite{ju2022prompting,nag2022zero,phan2024zeetad,narayan2021d2}, and few-shot \cite{nag2021few,lee2023few, thatipelli2022spatio} settings.
While open-set approaches typically assign an ``unknown’’ label to novel action categories, zero-shot methods rely on a text encoder's ability to provide meaningful representations based on the class name. However, the latter approaches have a tendency to overfit and are likely to be biased towards base categories.
Recent work \cite{phan2024zeetad} finetunes CLIP \cite{radford2021learning}, which comprises a vision and text encoder for encoding images and corresponding text labels. Although finetuning CLIP's text encoder helps bridge the domain gap between the videos and text in the downstream task, it comes at the cost of losing the generalization learned between the CLIP visual and text encoders. This is because only the text encoder is finetuned with fixed prompts involving only the class names for the downstream task.
In contrast, we propose to encode rich class-specific language descriptions (extracted from an LLM) using the CLIP text encoder and utilize them as
guidance features for learning the visual cues and semantic context related to novel action categories. Overall, our approach harnesses the power of LLMs and the internal representation of the CLIP text encoder to provide rich and informative descriptions for novel action categories. 

Language descriptions enable the ability to clearly distinguish between closely related actions having similar visual cues. For example, \texttt{javelin throw} and \texttt{pole vault} actions have visual similarities such as \texttt{sports fields}, \texttt{equipment}, and body motion such as \texttt{running}, \texttt{jumping} and \texttt{throwing}. 
To leverage these descriptions for localizing actions, we propose to learn multimodal guided features by first cross-attending the language descriptions with frame-level (spatial) features. These guided features are then fused with snippet-level (spatio-temporal) features to achieve multimodal snippet-level features. Such a progressive integration of language descriptions to spatio-temporal features through the spatial features achieves a better alignment between textual embeddings and visual action features. This alignment aids in correctly localizing the novel actions based on their descriptions during inference. Furthermore, we employ a two-stage training pipeline, in which we first train our proposed \ovformer on a larger vocabulary dataset, followed by finetuning it on the downstream data to adapt to its characteristics.
To the best of our knowledge, this is the first work on OVTAL.
We formulate a simple but strong solution by leveraging LLMs and crafting task-specific prompts as input to generate class-specific language descriptions. We introduce the modality mixer module for fusing class-specific language descriptions with frame-level features to yield multimodal guided features. These features help learn the mapping between text embeddings and the visual cues related to the action. When fused with snippet-level features, this mapping is transferred to recognize novel action categories. We conduct extensive experiments on two popular benchmarks and significantly outperform existing SOTA approaches on \thumos \cite{idrees2017thumos} and \anet \cite{caba2015activitynet} for both OVTAL and ZSTAL tasks.
\section{Related Work}
\label{sec:related_work} 
\vspace{-0.2cm}

\noindent \textbf{Temporal Action Localization (TAL):}~
Existing TAL methods fall into two categories: two-stage approaches, which involve proposal generation followed by classification (based on anchor windows \cite{buch2017sst,heilbron2016fast,escorcia2016daps}, action boundaries \cite{gong2020scale,lin2019bmn,lin2018bsn,liu2019multi,zhao2020bottom}, graphs \cite{bai2020boundary,xu2020g}, or transformers \cite{chang2022augmented,tan2021relaxed,ranasinghe2022self}), and single-stage approaches \cite{lin2021learning,zhang2022actionformer}, which are anchor-free and trained end-to-end. However, a key limitation of all current TAL methods is their closed-world assumption --- they require the same action categories, ranging from around 20 to 200, to be present both during training and inference, preventing generalization to novel action categories unseen during training. \\
\noindent \textbf{Zero-Shot Temporal Action Localization (ZSTAL):}~To address this limitation, ZSTAL aims to localize and recognize novel action categories in untrimmed videos unseen during training. Traditional zero-shot learning approaches transfer knowledge from ``seen'' to ``unseen'' classes through shared semantic embeddings or vision-language alignments. Prior works are classified into semantic embedding-based approaches such as ZSTAD \cite{zhang2020zstad}, TranZAD \cite{nag2023semantics}, and vision-language model-based approaches such as Efficient-Prompt \cite{ju2022prompting}, STALE \cite{nag2022zero}, and ZEETAD \cite{phan2024zeetad}. However, zero-shot methods still fall short of real-world applications, specifically because of the constraint of identifying ``unseen'' categories without prior knowledge and relying solely on the base categories. Building upon the limitations of TAL and ZSTAL, we introduce OVTAL, which lifts the restriction of defining ``unseen'' categories \emph{a priori}. \\ 
\noindent \textbf{Prompt-based techniques:}
Prompting refers to designing an instruction which, when passed through the pretrained language model, can guide the downstream task. Prompt-based learning techniques have been widely used in the NLP domain \cite{jiang2020can,liu2023pre}. CLIP \cite{radford2021learning} introduces prompt-based learning in image recognition tasks, where it shows learning relationships between vision-language models using large-scale image-text pairs. Methods like \cite{sun2022dualcoop,zhou2022conditional,zhou2022learning} introduced learnable vectors to the text encoder of CLIP for transfer learning to recognition tasks. We use action description-based prompting in this work to enable the localization of novel action classes in the open-vocabulary setting. \\
In summary, while previous works like Efficient-Prompt, STALE, and ZEETAD explore low-shot temporal action localization, to the best of our knowledge, our work is the first to investigate the open-vocabulary setting. Our proposed approach leverages pretraining on a larger localization vocabulary dataset, fusing visual features with text descriptions from a language model to obtain rich multimodal representations. This enables the model to capture visual cues and semantic context related to the actions, leading to improved performance on both base and novel actions.
\section{Open-Vocabulary Temporal Action localization}
\label{sec:method} 
\vspace{-0.2cm}

\begin{figure*}[t!]
    \centering
    \includegraphics[width=0.95\linewidth]{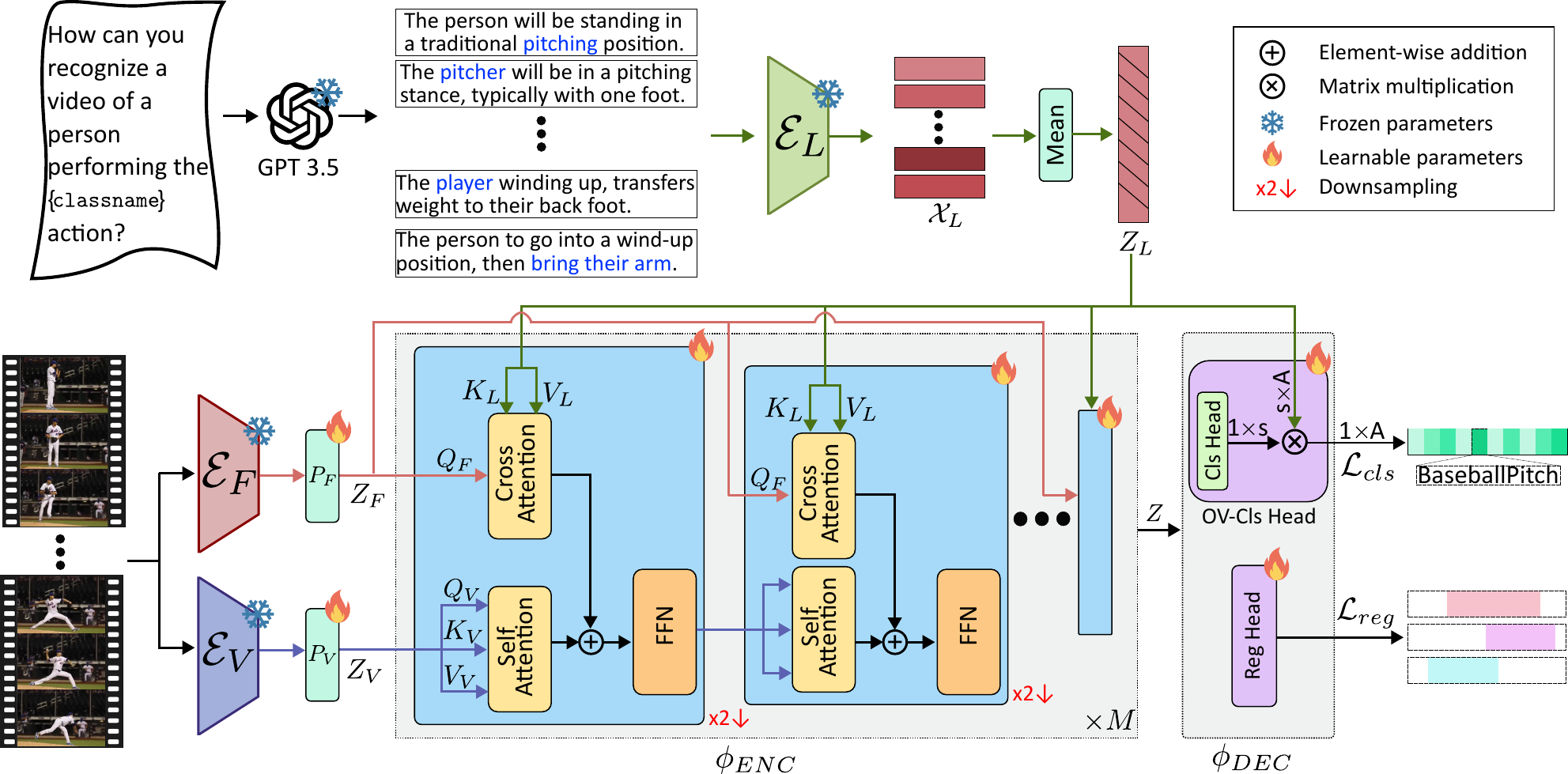}
    \vspace{0.3cm}
    \caption{\textbf{Overview of OVFormer.} Given a long untrimmed video $X$, frame- and snippet-level features are extracted and projected into $D$-dimensional feature spaces $Z_F$ and $Z_V$ using the projection functions $P_F$ and $P_V$, respectively.
    These features are then passed as input to the multi-scale $\phi_{ENC}$ module, which includes our proposed modality mixer. The modality mixer takes $Z_F$ and $Z_V$ as input, where $Z_V$ undergoes self-attention, and $Z_F$ is cross-attended with text embeddings $Z_L$ obtained from LLM-generated descriptions. The resulting multimodal guided features are fused with the self-attended $Z_V$. The output of $\phi_{ENC}$, enriched multimodal snippet-level features $Z$, is used as input for $\phi_{DEC}$, which consists of OV-classification and regression heads. The OV-classification head maps the enriched multimodal snippet-level features to the semantic space, relating them to class semantics and obtaining action candidates. During inference, text embeddings of novel categories are used to enable the OV capability.}
    \label{fig:main_arch}
    \vspace{-1.5em}
\end{figure*}



\noindent\textbf{Problem Formulation:}
\label{subsec:pf}
Given an input video $X$, frame-level features are denoted by $X_F = \{x_f^1, x_f^2, \cdots, x_f^T\}$ and snippet-level features by $X_V = \{x_v^1, x_v^2, \cdots, x_v^T\}$ over time $t = \{1, 2, \cdots, T\}$. Here, $T$ denotes the total duration of the video. When the feature vectors $\{x^t\}_{t=1}^T$ are fed as input to the 
OVTAL method,
the method is expected to predict action labels $Y = \{y_1, y_2, \cdots, y_N\}$, where $N$ is the number of action instances.
Each instance $y_i = \{s_i, e_i, a_i\}$ is defined by a start time $s_i$, end time $e_i$, and action label $a_i$, where $s_i \in [1,T]$, $e_i \in (s_i,T]$, and 
$a_i \in \{1,\cdots, A\}$, where $A$ is the number of action categories (elaborated on below).
Taking inspiration from \cite{zhou2022detecting,kaul2023multi}, two datasets are used during training: a large vocabulary-dense annotation dataset $\mathcal{V}_{super}$ with vocabulary $\mathcal{A}_{super}$, and a smaller dataset $\mathcal{V}_{base}$ with vocabulary $\mathcal{A}_{base}$. During inference, we use a testing split $\mathcal{V}_{novel}$ with vocabulary $\mathcal{A}_{novel}$ that shares the same data structure as $\mathcal{V}_{base}$.
To identify novel categories, text embeddings $Z_L$ are introduced into the training pipeline as input to
$\phi_{ENC}$ and $\phi_{DEC}$ to the OV-classification head.
In the most general case, there are no restrictions on the overlap or lack thereof between the sets $\mathcal{A}_{super}$, $\mathcal{A}_{base}$, and $\mathcal{A}_{novel}$.
In OVTAL, a deep network $f(\cdot)$ is trained to identify novel action categories.
The network is the composition of two modules $f = \phi_{DEC} \circ \phi_{ENC}$.
%
%
%
The encoder $\phi_{ENC}(X_V, X_F, Z_L)$ yields multi-scale representations $Z = \{Z^1, Z^2, ..., Z^M\}$, where $Z \in \mathbb{R}^{2^{m-1}T \times D}$ and $m = 1 \cdots M$. $Z$ are then passed through the decoder $\phi_{DEC}(\{Z^j\}^K_{j=1})$, which yields predicted labels 
$\hat{Y} = \{\hat{y}^1, \hat{y}^2, ..., \hat{y}^T\}$. In stage \RNum{1}, $f(\cdot)$ is trained on $\mathcal{V}{super}$ to learn from a larger vocabulary $\mathcal{A}{super}$ along with $Z_L$ class-specific language descriptions. This is followed by stage \RNum{2} training, where $f(\cdot)$, previously trained on $\mathcal{V}{super}$, is finetuned on $\mathcal{V}{Base}$ to adapt to the dataset characteristics of base action categories, resulting in improved performance.
Our goal for $f(\cdot)$ is to predict any action category from the combined set $\mathcal{A} = \mathcal{A}_{base} \cup \mathcal{A}_{novel}$ during inference. The proposed \ovformer aims to generalize effectively to novel action categories while maintaining high performance on base categories.

\subsection{Overall Architecture}
\label{subsec:overall_arch}

As previously discussed, localizing and recognizing novel action categories while remembering the base action categories is a challenging task. \autoref{fig:main_arch} shows the overall architecture of the proposed OVTAL method. \ovformer adapts the popular \aformer \cite{zhang2022actionformer} as its base architecture and introduces (i) class-specific language descriptions (\autoref{subsec:prompts}) from an LLM to classify and localize novel action categories; and (ii) a modality mixer (\autoref{subsec:modal_mixer}) for learning the scene information and semantic context by cross-attending aggregated text embeddings $Z_L$ and the frame-level features $Z_F$. Furthermore, by introducing $Z_L$ into the training pipeline, we are able to separate foreground action regions from the background and emphasize the visual cues and semantic context related to the actions.
In the proposed \ovformer, 
an input video $X$ is fed into modality-specific off-the-shelf encoders (video and visual) to obtain snippet- ($X_V$) and frame-level features ($X_F$).
These features are then passed through the projection functions $P_V$ and $P_F$ which embed them into $D$-dimensional space, $Z_V \in \mathbb{R}^{T \times D}$, and $Z_F \in \mathbb{R}^{T \times \hat{D}}$, respectively. Both of these are input to $\phi_{ENC}(\cdot)$ along with class-specific text embeddings $Z_L$. Here, $T$ is the temporal length, $D$ is the dimension of the feature vector for each snippet, and $\hat{D}$ is the dimension of the feature vector for each frame. $\phi_{ENC}$ captures multi-scale feature representations for frame-level and snippet-level features, i.e., $Z_F \in \mathbb{R}^{2^{m-1}T \times \hat{D}}$ and $Z_V \in \mathbb{R}^{2^{m-1}T \times D}$, where $m = 1  \cdots M$. These multi-scale representations, along with the class-specific text embeddings $Z_L \in \mathbb{R}^{s \times A}$, where $s$ is the text embedding dimension for each class, are fed into the modality mixer. The output from $\phi_{ENC}(\cdot)$ results in an enriched multimodal snippet-level features representation $Z \in \mathbb{R}^{2^{m-1}T \times D}$. The enriched features are then fed to $\phi_{DEC}(\cdot)$, which consists of OV-classification and regression heads. The OV-classification head feature space is mapped to the class-specific text embeddings $Z_L \in \mathbb{R}^{s \times A}$ to relate to the class semantics. Overall, our proposed OVFormer is trained end-to-end using dedicated classification ($\mathcal{L}_{cls}$) and regression ($\mathcal{L}_{reg}$) loss terms. Next, we present the \ovformer approach in detail.

\subsection{OVFormer}
\subsubsection{Class-Specific Language Descriptions}
\label{subsec:prompts}
Existing approaches, such as Efficient-Prompt \cite{ju2022prompting}, make use of simple prompts like \say{A video of \texttt{\{classname\}}} or \say{\texttt{\{classname\}}}. These methods rely on the strength of the text encoder to understand class attributes and information related to the class solely from the class name. However, such prompts are unable to highlight the important attributes and semantic context responsible for defining the action. This capability is crucial for localization and classification, as it helps to understand the scenes and background context for the action.
To this end, we leverage
a pretrained language model, specifically the \verb+GPT-3.5-turbo-instruct+ model from OpenAI. We generate 
10 detailed descriptions per class (\autoref{fig:main_arch} shows four descriptions for clarity, with more examples in Supplementary).
For generating rich, detailed descriptions of the class by LLM, we pass a prompt:
\say{How can you recognize a video of a person performing the \texttt{\{classname\}} action?}
Given a set of $E$ language descriptions ${s^a_r}$ for a predefined category $a$, we encode each description using the CLIP text encoder \cite{radford2021learning}, and obtain an aggregated embedding for the action category $a$ as:
\begin{equation}
Z_L = \frac{1}{E} \sum_{r=1}^{E} \mathcal{E}_L(s^a_r).\
\end{equation}
Using the aggregated embedding helps capture the semantics of the class while mitigating biases from individual descriptions. These 
embeddings $Z_L$ are used as input to the modality mixer and $\phi_{DEC}(\cdot)$ (as shown in \autoref{fig:main_arch}).
This simple technique of aggregation can summarize the class-wise description very well. During testing, $Z_L$ for novel actions are computed in the same way by passing novel categories as classnames to enable the OVTAL setting.

\subsubsection{Modality Mixer}
\label{subsec:modal_mixer}
A na\"ive approach for converting a fully-supervised TAL model to an OVTAL model is to simply multiply the classifier output features with textual features. However, such an approach is insufficient to handle novel action categories effectively since a late fusion of the two modalities likely results in the encoder learning less discriminative action features that are not well-aligned with the textual embeddings. Here, we strive to develop a more robust contextualization method for accurately detecting actions in untrimmed videos within an OVTAL setting. 
To this end, we introduce a modality mixer, a simple yet effective approach that enhances the 
snippet-level features $Z_V$ using textual embeddings $Z_L$ by capturing long-range temporal dependencies between the visual features and aligning them to the corresponding textual embeddings in a progressive manner, resulting in enriched multimodal snippet-level features $Z$.

Capturing long-range temporal dependencies is crucial in OVTAL, as actions may span across multiple time steps, and the context surrounding an action is likely to provide valuable information for accurate recognition and localization. Thus, our modality mixer first focuses on learning the temporal context across the full sequence. Here, the features $X_V$ are projected into $Z_V$ using a convolutional network consisting of two $1\times1$ convolution layers with ReLU, where $Z_V \in \mathbb{R}^{T \times D}$ with $T$ time steps and $D$ dimensional features. These features are projected into a low-dimensional space for creating query, key, and value tensors given by $Q_{V}^{h} = Z_{V}W_{Q}^{h}$, $K_{V}^{h} = Z_{V}W_{K}^{h}$ and $V_{V}^{h} = Z_{V}W_{V}^{h}$, which self-attend to result in enriched features $Z_{V}^{'}$ given by
\begin{equation}
Z_{V}^{'} = \left[ \alpha^1; \alpha^2; \ldots; \alpha^H \right]W_o, \quad \text{where} \quad \alpha^{h} = A^{h}V^{h} \quad   \text{with} \quad A^{h} = \sigma\left(\frac{Q_{V}^{h}(K_{V}^{h})^{T}}{\sqrt{D_{k}}}\right).
\label{eq:sa_block}
\end{equation}
%
%
Here, $h \in \{1, 2, \ldots, H\}$, and $W_{Q}^{h}, W_{K}^{h}, W_{V}^{h}, W_o$ are learnable parameters. 
Consequently, the enriched snippet-level features $Z_{V}^{'}$ can encode the long temporal context.
Furthermore, we propose to enhance the alignment between the text embeddings $Z_{L}$ and the snippet-level features $Z_{V}$ well before the classification stage in a progressive manner. First, we align the frame-level features $Z_{F}$ with $Z_{L}$ through cross-attention and then fuse the resulting features with the enriched snippet-level features. Such a text $\rightarrow$ image (frame-level) $\rightarrow$ video (snippet-level) progressive integration aids in better aligning the visual features to the corresponding textual embeddings. The fused features are then passed through a feed-forward network. 
%
The query, key, and value tensors $Q_{F}^{h} = Z_{F}\hat{W}_{Q}^{h}$, $K_{L}^{h} = Z_{L}\hat{W}_{K}^{h}$ and $V_{L}^{h} = Z_{L}\hat{W}_{V}^{h}$ are used to obtain multimodal guided features $Z_F^{'}$, similar to \autoref{eq:sa_block}.
Furthermore, the enriched multimodal snippet-level features are computed as 
\begin{equation}
Z = FFN(Z_{F}^{'} + Z_{V}^{'})
\end{equation}
By embedding class-specific language descriptions within the training pipeline at an earlier stage, we ensure that the snippet-level features are more closely aligned with the textual descriptions by the time they reach the classifier. This early fusion of modalities enables our model to effectively recognize and localize novel action categories in untrimmed videos.
\subsection{Training and Inference}
Our proposed OVFormer is trained end-to-end using the following joint loss formulation:
\begin{equation}
\mathcal{L} = \left( \mathcal{L}_{cls} + \lambda\mathcal{L}_{reg} \right)
\end{equation}
where $\mathcal{L}_{cls}$ and $\mathcal{L}_{reg}$ denote the loss terms for the OV-classification and regression heads, respectively. For $\mathcal{L}_{cls}$, we employ the standard focal loss \cite{lin2017focal} for $A$-way binary classification, while for $\mathcal{L}_{reg}$, we utilize the standard DIoU loss \cite{zheng2020distance} for regression. The weighting factor $\lambda_{reg}$ is set to a default value of 1. At inference time, the novel action categories are passed as classnames to the prompt, which leads to $A_{novel}$ predictions from the OV-classification head, followed by predicted regression ranges from the regression head.

\section{Experiments}
\label{sec:exper}
\vspace{-0.2cm}

We evaluate \ovformer on two datasets: \thumos \cite{idrees2017thumos} and \anet \cite{caba2015activitynet}.
Following other open-vocabulary \cite{zareian2021open,gu2021zero,zhong2022regionclip} and TAL methods \cite{lin2019bmn,zeng2019graph,zhang2022actionformer,cheng2022tallformer}, we report mean average precision over base ($mAP_{base}$), novel ($mAP_{novel}$), and all ($mAP_{all}$) action categories.
Snippet- and frame-level features are extracted using a two-stream I3D video encoder \cite{carreira2017quo} and DINOv2 \cite{oquab2023dinov2} respectively for HACS, \thumos and \anet. Additional details on the experimental setup are provided in the supplementary material.
 \begin{table}[!t]
  \begin{minipage}{0.4\linewidth}
   \centering \small
  \setlength{\tabcolsep}{9pt}
  \scalebox{0.72}{
  \begin{tabular}{c|cc}
  \toprule
  \rowcolor{Gray} 
    \textbf{Method}  & \textbf{\thumos} &\textbf{\anet} \\
    \midrule
   P-\aformer  & 0.2  & 0.1 \\
   \rowcolor{LightCyan}
   \ovformer (Ours)  & \textbf{12.6} & \textbf{\textbf{19.0}} \\
  \bottomrule
  \end{tabular}}
  \end{minipage}
   \hfill
  \begin{minipage}{0.50\linewidth}
  \caption{ 
  Average performance ($mAP_{all}$) of P-\aformer[\autoref{Fig:variations}(a)] and \ovformer, both trained in Stage \RNum{1}
  and tested on \thumos and \anet over all classes. 
  }
 \label{table:nofinetune}
  \end{minipage}
     \vspace{-0.5em}
\end{table}
\begin{table*}[t]

\begin{center}
\caption{\textbf{OVTAL results on \thumos and \anet.} Average performance (mAP) over [0.3:0.1:0.7] for \thumos and [0.5:0.05:0.95] for \anet. Our proposed method, \ovformer, achieves significant gains in mAP over base, novel, and all action categories for both 75-25 and 50-50 splits. For a fair comparison, we evaluate STALE$^{\dagger}$ and obtain results for base, novel, and all action categories. See \autoref{sec:results} for more details.}
\label{table:ovtal_sota}
\vspace{0.15cm}
\setlength{\tabcolsep}{5pt} 
\scalebox{0.75}{
\begin{tabular}{c | l | c c c | c c c }
\toprule[0.1em]
\rowcolor{Gray}
& & \multicolumn{3}{c|}{\textbf{\thumos}} & \multicolumn{3}{c}{\textbf{\anet}}\\
\rowcolor{Gray}
\multirow{-2}{*}{\textbf{Train-Test split}} &  \multirow{-2}{*}{\textbf{ Method }} & 
          $\textbf{mAP}_{\textbf{base}}$ & $\textbf{mAP}_{\textbf{novel}}$ & $\textbf{mAP}_{\textbf{all}}$ & $\textbf{mAP}_{\textbf{base}}$ & $\textbf{mAP}_{\textbf{novel}}$ & $\textbf{mAP}_{\textbf{all}}$  \\

\midrule[0.1em]
         \multirow{6}{*}{\makecell{75\% Seen \\ 25\% Unseen}}  & \aformer\cite{zhang2022actionformer} & 65.1 & - & - & 31.0 & - & - \\
\cline{3-8}
        & P-\aformer & 51.9 & 13.8 & 41.5  & 30.0 & 15.3 & 26.3  \\
        & L-\aformer  & 52.3 & 14.7 & 42.8 & 30.9 & 16.8 & 27.3  \\
          & F-\aformer & 50.8 & 24.2  & 44.1 & 30.8 & 22.9 & 28.8 \\
        & STALE{$^\dagger$} \cite{nag2022zero} & - & - & - & 23.2 & 20.6  & 22.6 \\
\cline{3-8}
          & \CC \ovformer (ours) & \CC \textbf{56.4} & \CC \textbf{27.3} & \CC \textbf{49.1} & \CC \textbf{31.4} & \CC \textbf{25.1} & \CC \textbf{29.8} \\

\midrule[0.1em]
         \multirow{6}{*}{\makecell{50\% Seen \\ 50\% Unseen}} & \aformer\cite{zhang2022actionformer} & 63.1 & - & - & 28.6 & - & -  \\
\cline{3-8}
        & P-\aformer & 50.9 & 9.9 & 30.5  & 27.6  & 13.0  & 20.3 \\
         & L-\aformer & 48.3 & 10.1 & 29.2 & 28.3 & 13.5 & 20.9  \\
          & F-\aformer & 51.2 & 20.5 & 35.8 & 28.8 & 23.5  & 26.2 \\
        & STALE{$^\dagger$} \cite{nag2022zero} & - & - & - & 23.0 & 20.7  & 22.2 \\
\cline{3-8}
          & \CC \ovformer (ours) & \CC \textbf{55.7} & \CC \textbf{24.9} & \CC \textbf{40.7} & \CC \textbf{30.2} & \CC \textbf{24.8} & \CC \textbf{27.5} \\
\bottomrule[0.1em]
\end{tabular}}
\end{center}
\vspace{-2em}
\end{table*}

\begin{figure*}[!t]
    \centering
    \includegraphics[width=1.0\linewidth]{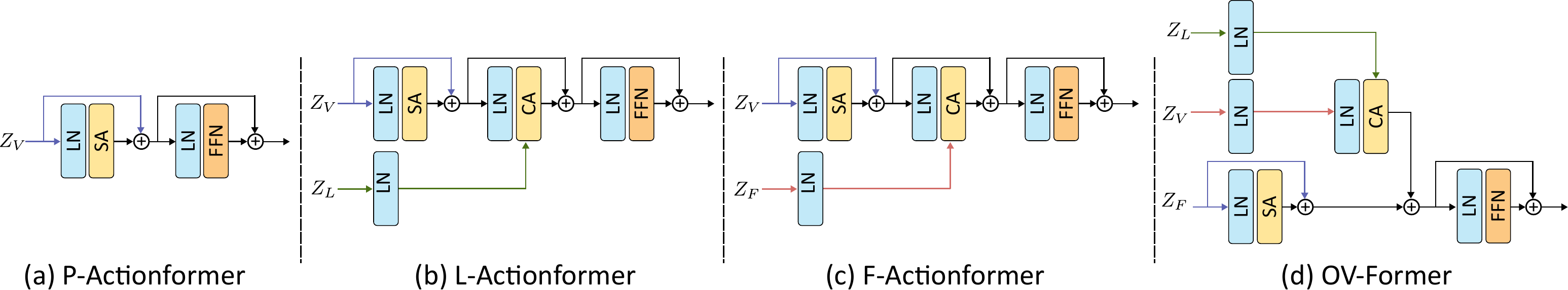}
    \vspace{0.15cm}
    \caption{Design choices for the modality mixer which are used as baselines for the OVTAL setting and evaluated in \autoref{table:ovtal_sota}. From (a-d) the text embeddings $Z_L$ are introduced in the OV-classification head (a) Na\"ive solution where only snippet-level features.
    (b) Introduce text embeddings and cross-attend with the snippet-level features.
    (c) A variation on (b) 
    where frame-level features are cross-attended with snippet-level features. (d) Our proposed method cross-attends text embeddings with frame-level features to learn multimodal guided features, which is fused with snippet-level features.}
    \label{Fig:variations}
    \vspace{-1em}
\end{figure*}

\subsection{Results}
\label{sec:results}

As this is the first exploration of Open-Vocabulary in TAL, we study three baselines based on our \ovformer: P-\aformer, L-\aformer, and F-\aformer (\autoref{Fig:variations}(a)-(c), respectively) and compare their performances to that of our OVFormer model (\autoref{Fig:variations}(d)).

\noindent \textbf{Pretraining generalization:} In \autoref{table:nofinetune}, \ovformer and P-\aformer models with Stage \RNum{1} training alone are directly evaluated on \thumos and \anet, illustrating the outcomes (i) when only Stage \RNum{1} is used without Stage \RNum{2} and (ii) the effect of fusing text embeddings at the classifier. The baseline (P-\aformer), which introduces text embeddings \textit{only} in the OV-classification head, performs poorly on novel action categories (0.2\% mAP on \thumos, 0.1\% on \anet). This indicates that late fusion of text embeddings is insufficient to localize and recognize novel action categories and Stage \RNum{1} alone is insufficient to bridge the gap between datasets with different characteristics. In contrast, our proposed method introduces text embeddings in the training pipeline and fuses them with snippet-level features, focusing on learning scene information and semantic context. This helps to separate foreground and background objects, leading to improved generalization performance on novel categories (12.6\% mAP on \thumos, 19.0\% on \anet).

\begin{table*}[!t]
\centering
\caption{\textbf{State-of-the-art comparison for ZSTAL on \thumos and \anet.} We show the comparison in terms of mAP evaluated over novel action categories and IoU thresholds of [0.3:0.1:0.7] for \thumos and [0.5:0.05:0.95] for \anet. Our \ovformer achieved significant gains in mAP in comparison to existing approaches. We only include the methods with open-source code available. See sec. \autoref{sec:results} for more details.}
\vspace{0.15cm}
\setlength{\tabcolsep}{6pt} 
\scalebox{0.75}{
\begin{tabular}{c| l | c c c c c c| c c c c c}
\toprule[0.1em]
\rowcolor{Gray}
 &  & \multicolumn{6}{c|}{\textbf{\thumos}} & \multicolumn{4}{c}{\textbf{\anet}} \\
\rowcolor{Gray}
\multirow{-2}{*}{\textbf{Train-Test split}} &  \multirow{-2}{*}{\textbf{ Method }} & 
           0.3  & 0.4  & 0.5  & 0.6  & 0.7 & \textbf{mAP} & 0.5  & 0.75  & 0.95  & \textbf{mAP}\\
\midrule[0.1em]
 \multirow{5}{*}{\makecell{75\% Seen \\ 25\% Unseen}} & B-II \cite{nag2022zero} & 28.5 & 20.3 & 17.1 & 10.5 & 6.9 & 16.6 & 32.6 & 18.5 & 5.8   & 19.6 \\
           & B-I \cite{nag2022zero} & 33.0 & 25.5 & 18.3 & 11.6 & 5.7 & 18.8 & 35.6 & 20.4 & 2.1 & 20.2 \\
           & Eff-Prompt \cite{ju2022prompting} & 39.7 & 31.6 &23.0 & 14.9 & 7.5 & 23.3 & 37.6 & 22.9 & 3.8 & 23.1 \\
           & STALE \cite{nag2022zero} & 40.5 & 32.3 & 23.5 & 15.3 & 7.6 & 23.8 & 38.2 & 25.2 & 6.0 & 24.9 \\
           & \CC \ovformer (ours) & \CC \textbf{49.8} & \CC \textbf{43.8} & \CC \textbf{35.8} & \CC \textbf{27.8} & \CC \textbf{19.2} & \CC \textbf{35.3}{\tiny{\textcolor{teal}{$\mathord{\uparrow}11.5$}}} & \CC \textbf{46.7} & \CC \textbf{29.4} & \CC \textbf{6.1} & \CC \textbf{29.5}{\tiny{\textcolor{teal}{$\mathord{\uparrow}4.6$}}} \\
\midrule[0.1em]
\multirow{5}{*}{\makecell{50\% Seen \\ 50\% Unseen}} & B-II \cite{nag2022zero} & 21.0 & 16.4 & 11.2 & 6.3 & 3.2 & 11.6 & 25.3 & 13.0 & 3.7 & 12.9 \\
           & B-I \cite{nag2022zero} & 27.2 & 21.3 & 15.3 & 9.7 & 4.8 & 15.7 & 28.0 & 16.4 & 1.2 & 16.0 \\
           & Eff-Prompt \cite{ju2022prompting} & 37.2 & 29.6 & 21.6 & 14.0 & 7.2 & 21.9 & 32.0 & 19.3 & 2.9 & 19.6 \\
           & STALE \cite{nag2022zero} & 38.3 & 30.7 & 21.2 & 13.8 & 7.0 & 22.2 & 32.1 & 20.7 & 5.9 & 20.5 \\
           & \CC \ovformer (ours) & \CC \textbf{42.8} & \CC \textbf{37.3} & \CC \textbf{30.6} & \CC  \textbf{23.5} & \CC \textbf{15.9} & \CC \textbf{30.5}{\tiny{\textcolor{teal}{$\mathord{\uparrow}8.3$}}} & \CC \textbf{42.8} & \CC \textbf{27.3} & \CC \textbf{6.0} & \CC \textbf{27.2} {\tiny{\textcolor{teal}{$\mathord{\uparrow}6.7$}}} \\
\bottomrule[0.1em]
\end{tabular}}
\label{tab:zstad_sota}
\vspace{-0.5em}
\end{table*}
\noindent \textbf{Performance on OVTAL:} \autoref{table:ovtal_sota} shows the state-of-the-art performance on the OVTAL task. We report results for our proposed \ovformer as well as the standard \aformer \cite{zhang2022actionformer} for comparison. Since \aformer can only localize and recognize base action categories, it is not directly applicable to OVTAL, and its $mAP_{novel}$ cannot be computed. For a fair comparison with an existing ZSTAL approach, we extended STALE$^\dagger$ \cite{nag2022zero} to
get $mAP_{base}$, $mAP_{novel}$, and $mAP_{all}$ scores. STALE$^\dagger$ achieves 23.2\%, 20.6\%, and 22.6\% for base, novel, and all categories, respectively. Our \ovformer significantly outperforms STALE, achieving 31.4\%, 25.1\%, and 29.8\% for the same categories.
The consistent performance gains across both the \thumos and \anet datasets highlight the effectiveness of our proposed contributions for the OVTAL task.

\noindent \textbf{Comparison of ZSTAL Methods:} We present a performance comparison for the ZSTAL task in \autoref{tab:zstad_sota}. We compared our method only with those that have available open-source implementations. Our \ovformer achieves significant improvements on both the \thumos and \anet. Following \cite{nag2022zero,ju2022prompting}, the evaluation is performed by considering only novel action categories during inference for the 75-25 and 50-50 splits. \ovformer outperforms existing ZSTAL methods by a substantial margin, illustrating the benefits of learning on a large vocabulary dataset and effectively modelling rich scene information.
\begin{figure*}[!t]
    \centering
    \includegraphics[width=1.0\linewidth]{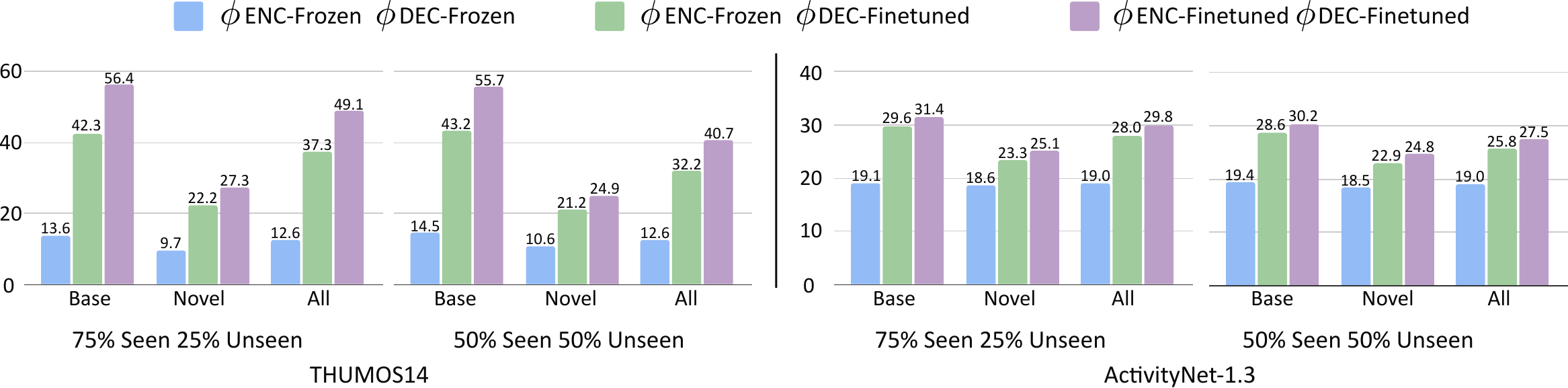}
    \vspace{0.15cm}
    \caption{\small \textbf{Finetuning strategies} by freezing or finetuning the $\phi_{ENC}$/$\phi_{DEC}$ on OVTAL setting. Here, for showing the effectiveness of Stage \RNum{2}, Stage \RNum{1} of the training pipeline is always present.}
    \label{fig:ovtal_finetuning}
    \vspace{-1.5em}
\end{figure*}

\subsection{Ablation Study}
\label{sec:ablation}
\autoref{fig:ovtal_finetuning} shows different finetuning strategies for Stage \RNum{2} on downstream data, where we observed that finetuning both $\phi_{ENC}$ and $\phi_{DEC}$ in our proposed method helps maintain overall performance while mitigating performance degradation on novel action categories.

\begin{figure*}[ht]
\centering
\begin{minipage}{0.59\linewidth}
\centering
\includegraphics[width=\linewidth]{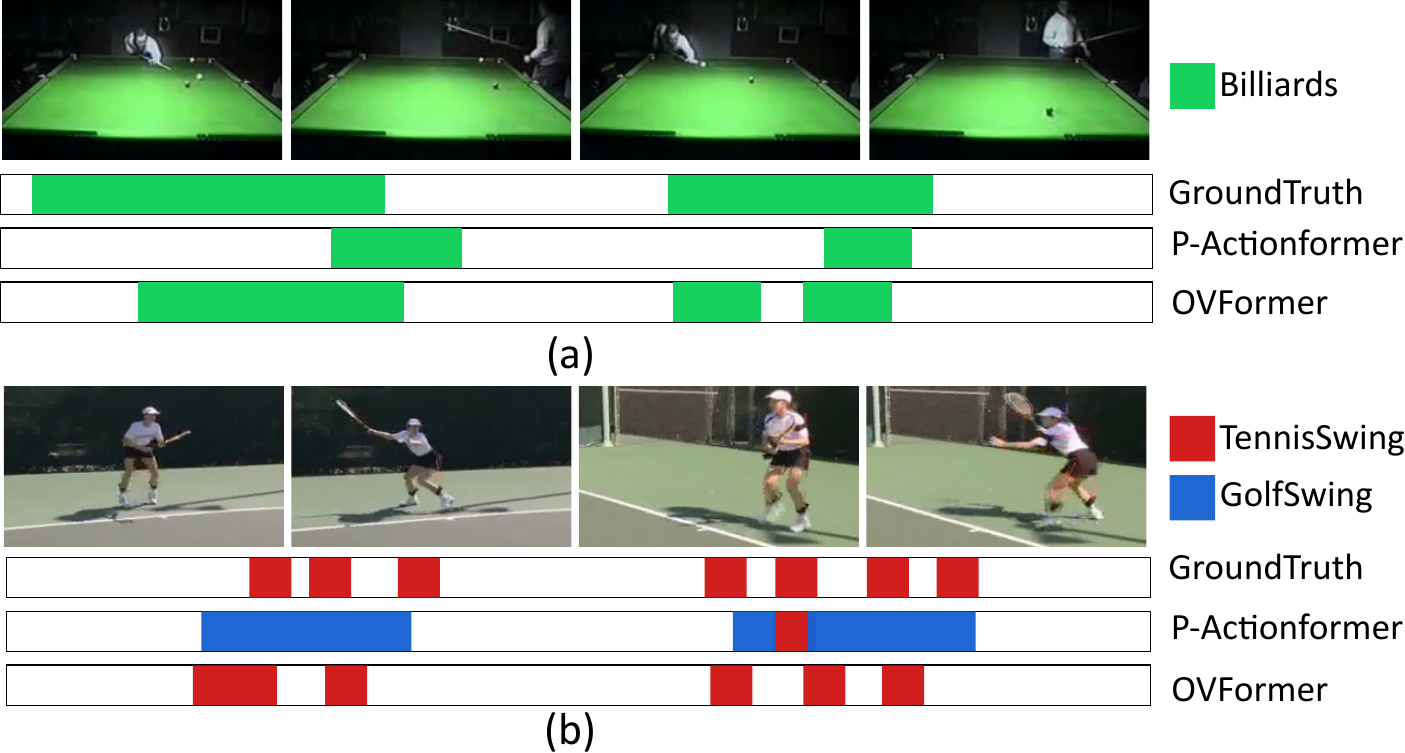}
\end{minipage}
\hfill
\begin{minipage}{0.38\linewidth}
\caption{\ovformer performance on \thumos in the OVTAL setting. We compare the performance of P-\aformer (\autoref{Fig:variations}(a)) and \ovformer (\autoref{Fig:variations}(d)) on (a) the \texttt{billiards} action, and (b) the \texttt{tennis swing} and \texttt{golf swing} actions.
}
\label{Fig:tal_qual}
\end{minipage}
\vspace{-0.5em}
\end{figure*}

\autoref{Fig:tal_qual} demonstrates the superior capabilities of \ovformer over P-\aformer. Our model predictions closely align with the ground truth, particularly in \texttt{billiards} and \texttt{tennis swing}. We examine the performance on \autoref{Fig:tal_qual}(a) \texttt{billiards}, and \autoref{Fig:tal_qual}(b) \texttt{tennis swing} and \texttt{golf swing} actions. In \autoref{Fig:tal_qual}(b), \texttt{tennis swing} belongs to the base classes, while \texttt{golf swing} belongs to the novel classes. In the case of P-\aformer, confusion exists between these actions, as they both have similar visual cues, \textit{i.e.}, a person running with an object in their hands. \ovformer improves scene information and semantic context by obtaining multimodal guided features and fusing them with snippet-level features, enhancing the separation between base and novel actions.

\section{Conclusions}
\label{sec: concl}

\vspace{-0.2cm}

In this work, we introduced Open-Vocabulary Temporal Action Localization, a novel and challenging task that aims to localize and recognize both base and novel action classes in untrimmed videos. To address this task, we proposed \ovformer, a framework that leverages multimodal guided features to enrich snippet-level features.
Our two-stage training strategy, which includes pretraining on a larger vocabulary dataset and finetuning on the downstream data, enables OVFormer to achieve state-of-the-art performance on both \thumos and \anet. The proposed approach significantly outperforms existing methods in both the OVTAL and ZSTAL settings, demonstrating its effectiveness in recognizing and localizing novel action categories while maintaining high performance on base categories. 


\section{Acknowledgment}

Resources used in preparing this research were provided, in part, by the Province of Ontario, the Government of Canada through CIFAR, and  \href{http://www.vectorinstitute.ai/partnerships/current-partners/}{partners of the Vector Institute}. GWT acknowledges support from NSERC. 

\bibliography{egbib}

\appendix

\section{Supplementary material}

\makeatletter
\renewcommand \thesection{A\@arabic\c@section}
\renewcommand \thesubsection{A\@arabic\c@section.\@arabic\c@subsection}
\renewcommand \thesubsubsection{A\@arabic\c@section.\@arabic\c@subsection.\@arabic\c@subsubsection}
\renewcommand\thetable{A\@arabic\c@table}
\renewcommand \thefigure{A\@arabic\c@figure}
\makeatother

In this supplementary material, we provide additional quantitative and qualitative analysis of our proposed Open-Vocabulary Temporal Action Localization (OVTAL) framework, \ovformer. Additional implementation details and quantitative results are discussed in \autoref{sec:suppl_impl}, \autoref{sec:suppl_exp}, followed by qualitative analysis in \autoref{sec:suppl_qual}. Finally, we provide details for the LLM-generated text descriptions for \thumos (\autoref{sec:suppl_gentext_th}) and \anet (\autoref{sec:suppl_gentext_anet}) used in the main manuscript. \\

\section{Additional Implementation details\label{sec:suppl_impl}}

\noindent \textbf{Datasets:}~We evaluate \ovformer on two datasets: \thumos \cite{idrees2017thumos} and \anet \cite{caba2015activitynet}.
\thumos consists of 20 classes and contains 413 untrimmed videos, while Activi-Net-1.3 is a large-scale dataset with 200 classes and 14,950 videos. Following \cite{ju2022prompting}, we divide the datasets into training and testing sets. Furthermore, we consider two settings: (A) training on 75\% of the action categories and testing on the remaining 25\%, and (B) training on 50\% of the categories and testing on the other 50\%. For \thumos, setting (A) involves 15 categories for training and 5 for testing, whereas setting (B) uses 10 categories for both training and testing. For \anet, setting (A) assigns 150 categories for training and 50 for testing, while setting (B) uses 100 categories for both training and testing. In each setting, we randomly sample the categories 10 times to create training and testing splits, and we report the average performance across these splits. 
For pretraining, we utilize the HACS dataset \cite{zhao2019hacs}, a large-scale dataset with dense annotations. Importantly, the HACS OV split, consisting of 24,407 videos, does not overlap with the testing splits of \thumos and \anet, ensuring a fair evaluation of \ovformer generalization capabilities. \\

\noindent \textbf{Evaluation Metrics:}~Following other image-based open-vocabulary approaches \cite{zareian2021open,gu2021zero,zhong2022regionclip} and TAL methods \cite{lin2019bmn,zeng2019graph,zhang2022actionformer,cheng2022tallformer}, we report mean average precision over base ($mAP_{base}$), novel ($mAP_{novel}$), and all ($mAP_{all}$) categories. The $mAP_{all}$ is used to show the model's performance across all action classes when both base and novel categories are present during inference. 
The $mAP_{all}$ is the most important metric: achieving a balance between $mAP_{base}$ and $mAP_{novel}$ is important, and while improving $mAP_{novel}$, a model should not improve $mAP_{novel}$ at the cost of degrading $mAP_{base}$. 
For ZSTAL \cite{ju2022prompting,nag2022zero}, we report mAP averaged over novel action categories. \\

\noindent \textbf{Implementation Details:} ~Our  architecture is based on \aformer \cite{zhang2022actionformer}. 
Frame-level features and snippet-level features are extracted using DINOv2 \cite{oquab2023dinov2} and a two-stream I3D video encoder \cite{carreira2017quo} for HACS, \thumos and \anet datasets. 
For pretraining using the HACS dataset, we use a temporal length of 512, a learning rate of $1e-3$, 40 epochs, and an NMS threshold of 0.75. Furthermore, for finetuning with THUMOS14, we use a temporal length of 2304, a learning rate of $1e-4$, 13 epochs, and an NMS threshold of 0.5. Similarly, for finetuning with ActivityNet-v1.3, we use a temporal length of 192, a learning rate of $1e-3$, 15 epochs, and an NMS threshold of 0.7. 
To generate text descriptions, we use the \texttt{gpt-3.5-turbo-instruct} model available from OpenAI and compute the text embedding using the CLIP ViT-B/32 text encoder model \cite{radford2021learning}. All experiments are performed using a single NVIDIA A100 GPU.

\begin{table*}[]
\begin{center}
\caption{
\textbf{Effect of different prompt templates on OVTAL setting for \ovformer.} Using our rich LLM-generated class-specific language descriptions during training to obtain multimodal guided features for the snippet-level features improves the $mAP_{novel}$ performance compared to manually crafted prompts.}
\vspace{0.3cm}
\label{table:ovtal_abl_prompt_tem}
\setlength{\tabcolsep}{5pt} 
\scalebox{0.65}{
\begin{tabular}{c | l |  c c c | c c c }
\toprule[0.1em]
\rowcolor{Gray}
& & \multicolumn{3}{c|}{\textbf{\thumos}} & \multicolumn{3}{c}{\textbf{\anet}}\\
\rowcolor{Gray}
\multirow{-2}{*}{\textbf{Split}} & \multirow{-2}{*}{\textbf{Prompt}}
         & \textcolor{gray}{$\textbf{mAP}_{\textbf{base}}$} & $\textbf{mAP}_{\textbf{novel}}$ & \textcolor{gray}{$\textbf{mAP}_{\textbf{all}}$} & \textcolor{gray}{$\textbf{mAP}_{\textbf{base}}$} & $\textbf{mAP}_{\textbf{novel}}$ & \textcolor{gray}{$\textbf{mAP}_{\textbf{all}}$}  \\

\midrule[0.1em]
          \multirow{3}{*}{\makecell{75\% Seen \\ 25\% Unseen}} 
          & \texttt{\{classname\}}  & \textcolor{gray}{59.3} & 8.0 & \textcolor{gray}{46.3}  & \textcolor{gray}{28.6} & 8.1  & \textcolor{gray}{23.6} \\
          & A video of \texttt{\{classname\}}  & \textcolor{gray}{59.2} & 8.5 & \textcolor{gray}{46.5}  & \textcolor{gray}{28.4} & 6.1 & \textcolor{gray}{22.8} \\
\cline{3-8}
          & \CC Ours: LLM generated descriptions & \CC \textcolor{gray}{59.0} & \CC \textbf{10.2}  & \CC \textcolor{gray}{46.8} & \CC \textcolor{gray}{28.7} & \CC \textbf{9.5}  & \CC \textcolor{gray}{23.9} \\
\midrule[0.1em]
          \multirow{3}{*}{\makecell{50\% Seen \\ 50\% Unseen}}
          & \texttt{\{classname\}} & \textcolor{gray}{59.0} & 6.1 & \textcolor{gray}{32.4} & \textcolor{gray}{26.2} & 5.1 & \textcolor{gray}{15.8} \\
          & A video of \texttt{\{classname\}}  & \textcolor{gray}{58.9}  & 7.0  &  \textcolor{gray}{32.8}  & \textcolor{gray}{25.9} & 4.3 & \textcolor{gray}{15.1} \\
\cline{3-8}
          & \CC Ours: LLM generated descriptions & \CC \textcolor{gray}{58.4} & \CC \textbf{7.7} & \CC \textcolor{gray}{33.1} & \CC \textcolor{gray}{26.2}  & \CC \textbf{6.8} & \CC \textcolor{gray}{16.5} \\
\bottomrule[0.1em]
\end{tabular}}
\end{center}
\end{table*}

\begin{table*}[]
\begin{minipage}{0.50\textwidth}
\setlength{\tabcolsep}{4pt}
\scalebox{0.50}{
\begin{tabular}{c | l | c c c}
\toprule[0.1em]
\rowcolor{Gray}
& & \multicolumn{3}{c}{\textbf{\thumos}} \\
\rowcolor{Gray}
\multirow{-2}{*}{\textbf{Train-Test Split}} & \multirow{-2}{*}{\textbf{Visual Encoder ($\mathcal{E_F}$) }}
         & $\textbf{mAP}_{\textbf{base}}$ & $\textbf{mAP}_{\textbf{novel}}$ & $\textbf{mAP}_{\textbf{all}}$ \\
\midrule[0.1em]
          \multirow{2}{*}{\makecell{75\% Seen \\ 25\% Unseen}} 
          & CLIP  & 50.5 & 21.1 & 43.2 \\
          & \CC DINOv2 & \CC \textbf{56.4} & \CC \textbf{27.3}  & \CC \textbf{49.1}\\
\midrule[0.1em]
          \multirow{2}{*}{\makecell{50\% Seen \\ 50\% Unseen}}
          & CLIP  & 50.3 & 17.1 & 33.7\\
          & \CC DINOv2 & \CC \textbf{55.7} & \CC \textbf{24.9} & \CC \textbf{40.7}\\
\bottomrule[0.1em]
\end{tabular}}
\end{minipage}%
\hfill
\begin{minipage}{0.50\textwidth}
\caption{\small \textbf{OVTAL results on THUMOS14}. Average performance (mAP) over [0.3:0.1:0.7] for \thumos and [0.5:0.05:0.95] for \anet. Our proposed method \ovformer using DINOv2 as off-the-shelf visual encoder $\mathcal{E_F}$ for frame-level features $X_F$ achieves significant improvement over CLIP. More details in \autoref{sec:suppl_ef}.}
\label{tab:thumos_results_ff}
\end{minipage}
\end{table*}


\section{Additional Quantitative Results\label{sec:suppl_exp}}

\subsection{Effect of different prompt templates} \autoref{table:ovtal_abl_prompt_tem} shows the \ovformer performance on manually crafted prompts and our class-specific generated descriptions from an LLM. Here, we demonstrate the performance using only Stage \RNum{2} of the training pipeline, without using additional data. We observe that using the simplest prompts, \say{\texttt{\{classname\}}} and \say{A video of \texttt{\{classname\}}}, achieves comparable performance to 
LLM-generated prompts for $mAP_{base}$ and $mAP_{all}$ but lower performance on $mAP_{novel}$. This demonstrates the importance of capturing the attributes and scene information surrounding the action. Using our proposed generated descriptions, we achieve improvement of 2.2\%, 1.4\%, 1.6\% and 1.7\% over 75-25 and 50-50 splits, respectively, for $mAP_{novel}$ compared to manually crafted prompts.

\subsection{Effect of Frame-Level Features \label{sec:suppl_ef}}

\autoref{tab:thumos_results_ff} presents the performance of our proposed method, \ovformer, using CLIP \cite{radford2021learning} and DINOv2 \cite{oquab2023dinov2} visual encoders $\mathcal{E_F}$ for extracting frame-level features $X_F$. We observe that off-the-shelf DINOv2 features significantly outperform CLIP features, with absolute gains of 5.9\%, 6.2\%, and 5.9\% over the 75-25 split for base, novel, and all action categories, respectively. Similarly, on the 50-50 split, DINOv2 achieves improvements of 5.4\%, 7.8\%, and 7.0\% over CLIP for the same categories. These results are consistent with the findings reported in \cite{oquab2023dinov2}, where DINOv2 is shown to capture richer visual descriptions compared to CLIP. This is particularly important for our problem statement, which focuses on body movements for related actions, as DINOv2's ability to capture richer visual descriptions helps in accurately distinguishing subtle differences in these movements. In this setup, both Stage \RNum{1} and Stage \RNum{2} of our method are utilized.


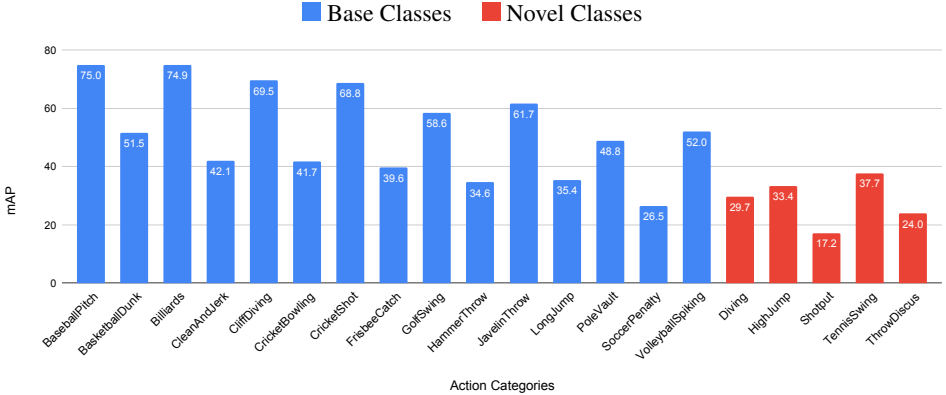
\begin{figure*}[t]
    \centering
    \small{
    \begin{tikzpicture}
\fill[color1bg] (0,0) rectangle (0.25,0.25);
\end{tikzpicture} Base Classes
\quad
    \begin{tikzpicture}
\fill[color2bg] (0,0) rectangle (0.25,0.25);
\end{tikzpicture} Novel Classes}
    \includegraphics[width=1.0\linewidth]{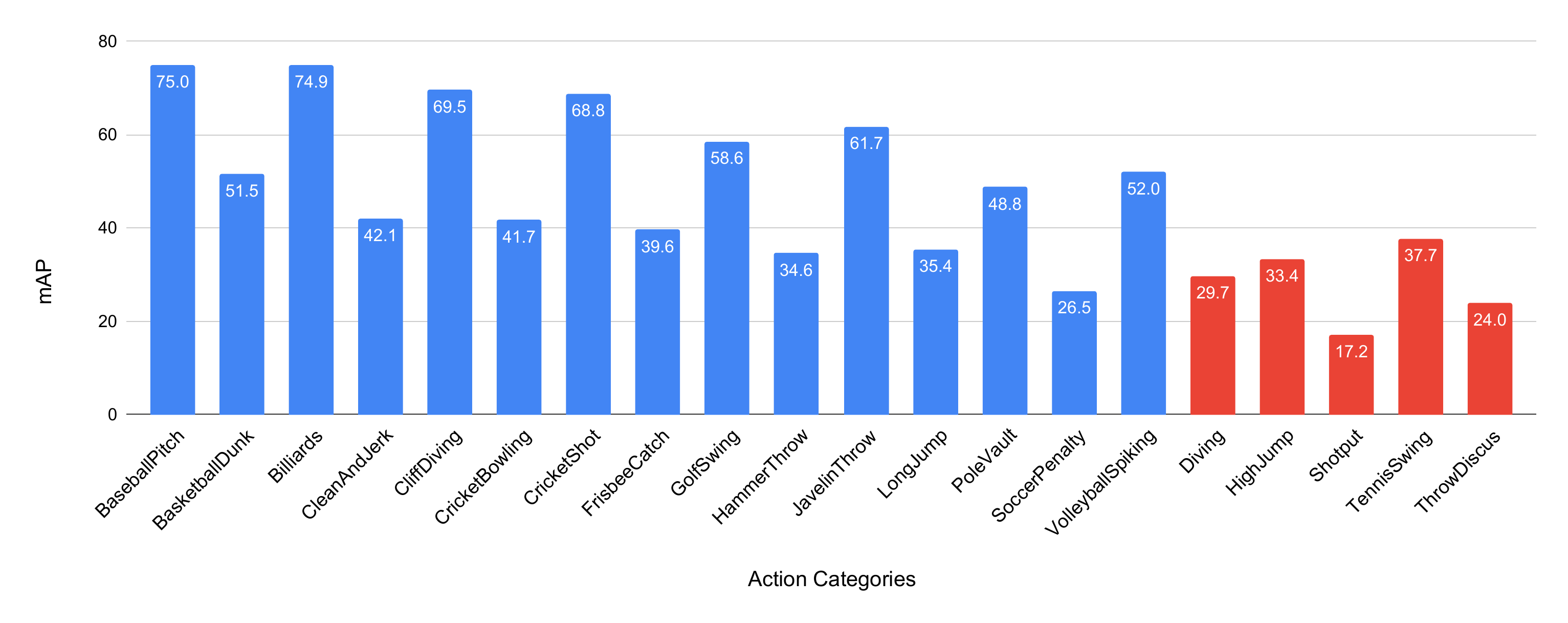}
    \vspace{0.1cm}
    \caption{\small Class-wise average $mAP$ for \thumos for 75-25 train-test split. }
    \label{fig:thumos_classwise7525}
\end{figure*}

\begin{figure*}[t]
    \centering
    \small{
    \begin{tikzpicture}
\fill[color1bg] (0,0) rectangle (0.25,0.25);
\end{tikzpicture} Base Classes
\quad
    \begin{tikzpicture}
\fill[color2bg] (0,0) rectangle (0.25,0.25);
\end{tikzpicture} Novel Classes}
    \includegraphics[width=1.0\linewidth]{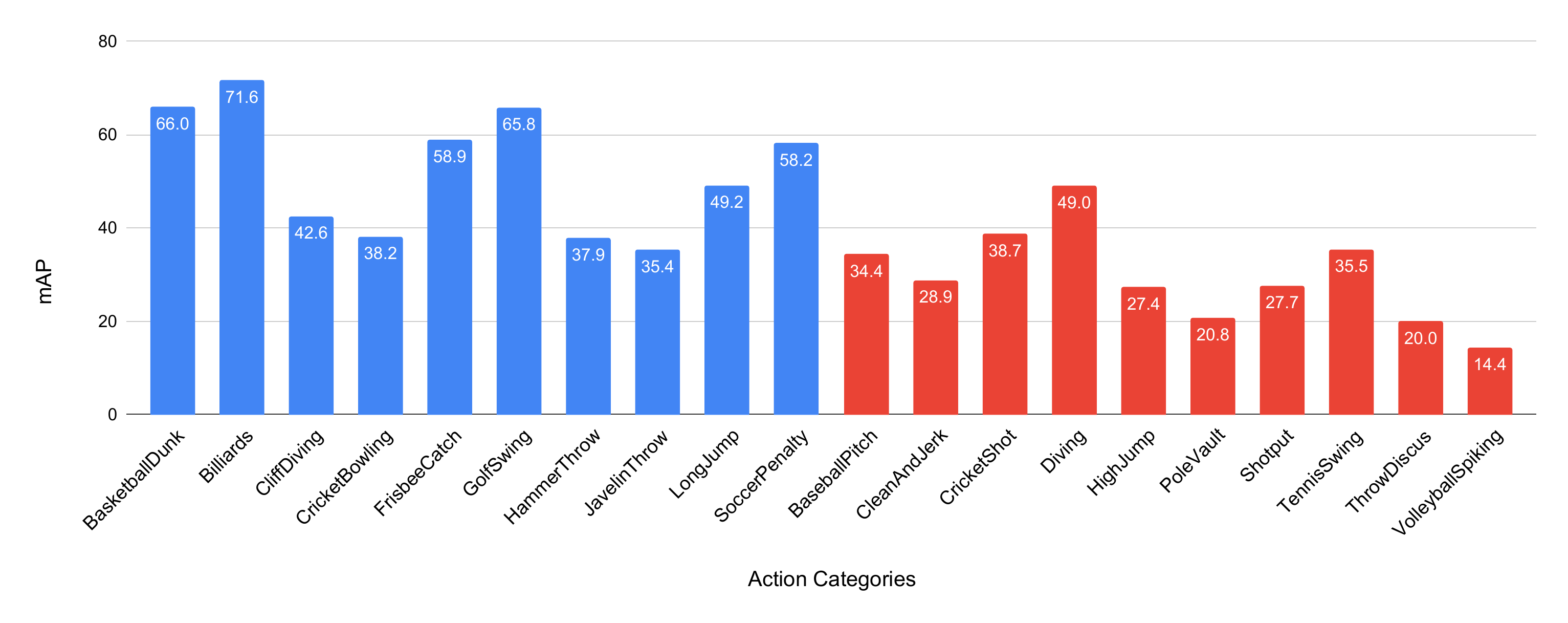}
    \vspace{0.1cm}
    \caption{\small Class-wise average $mAP$ for \thumos for 50-50 train-test split.}
    \label{fig:thumos_classwise5050}
\end{figure*}
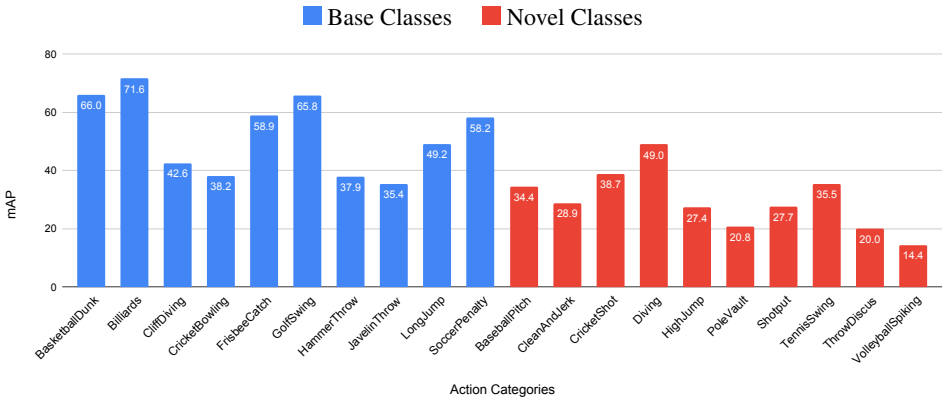

\subsection{Class-wise Average mAP}

In \autoref{fig:thumos_classwise7525} and \autoref{fig:thumos_classwise5050}, we report class-wise results of \ovformer on \thumos for one of the 10 random splits \cite{ju2022prompting} on 75-25 and 50-50 train-test splits, respectively. Both plots show a high variance in average mAP among the classes, specifically for actions with very similar visual cues. For example, \texttt{HammerThrow} and \texttt{JavelinThrow} have mAP values of 37.9\% and 35.4\%, respectively, for the 50-50 split, while \texttt{FrisbeeCatch} and \texttt{CricketBowling} have mAP values of 39.6\% and 41.7\%, respectively, for the 75-25 split. We attribute this variance in mAP to the similarity in visual cues and body movements between these actions. For instance, a person in a throwing motion is a common visual cue shared by both \texttt{HammerThrow} and \texttt{JavelinThrow}. The similarity between these actions motivated us to incorporate rich class-specific language descriptions and integrate the learning of these descriptions alongside the snippet-level features in the form of multimodal guided features. Also, incorporating Stage \RNum{1} training aids in mitigating the issue of overfitting on the base dataset $\mathcal{V}_{base}$. As a result, our approach learns to distinguish these close similarities between fine-grained actions better and enhances the detection of novel action categories without overfitting on the base action categories. 

Our \ovformer achieves higher mAP values for the base action categories (shown in blue) compared to the novel ones (shown in red). This is expected, as the model has been trained on the base categories and can better recognize them during inference. However, \ovformer is able to maintain a reasonable performance on the novel action categories. The effectiveness of this method can be observed in the performance on novel action categories. 
For instance, in the 75-25 split, the novel action categories such as \texttt{Diving}, \texttt{HighJump}, \texttt{Shotput}, \texttt{TennisSwing}, and \texttt{ThrowDiscus} have mAP values ranging from 17.2\% to 37.7\%. Similarly, in the 50-50 split, the novel action categories have mAP values ranging from 14.4\% to 49.0\%. These results demonstrate that OVFormer can effectively generalize to unseen action categories by incorporating rich class-specific language descriptions and the multimodal guided features. \ovformer is able to better distinguish between visually similar actions and improve performance on novel action categories that were not seen during training.

\section{Additional Qualitative Results \label{sec:suppl_qual}}

\begin{figure*}[t]
    \centering
    \includegraphics[width=1.0\linewidth]{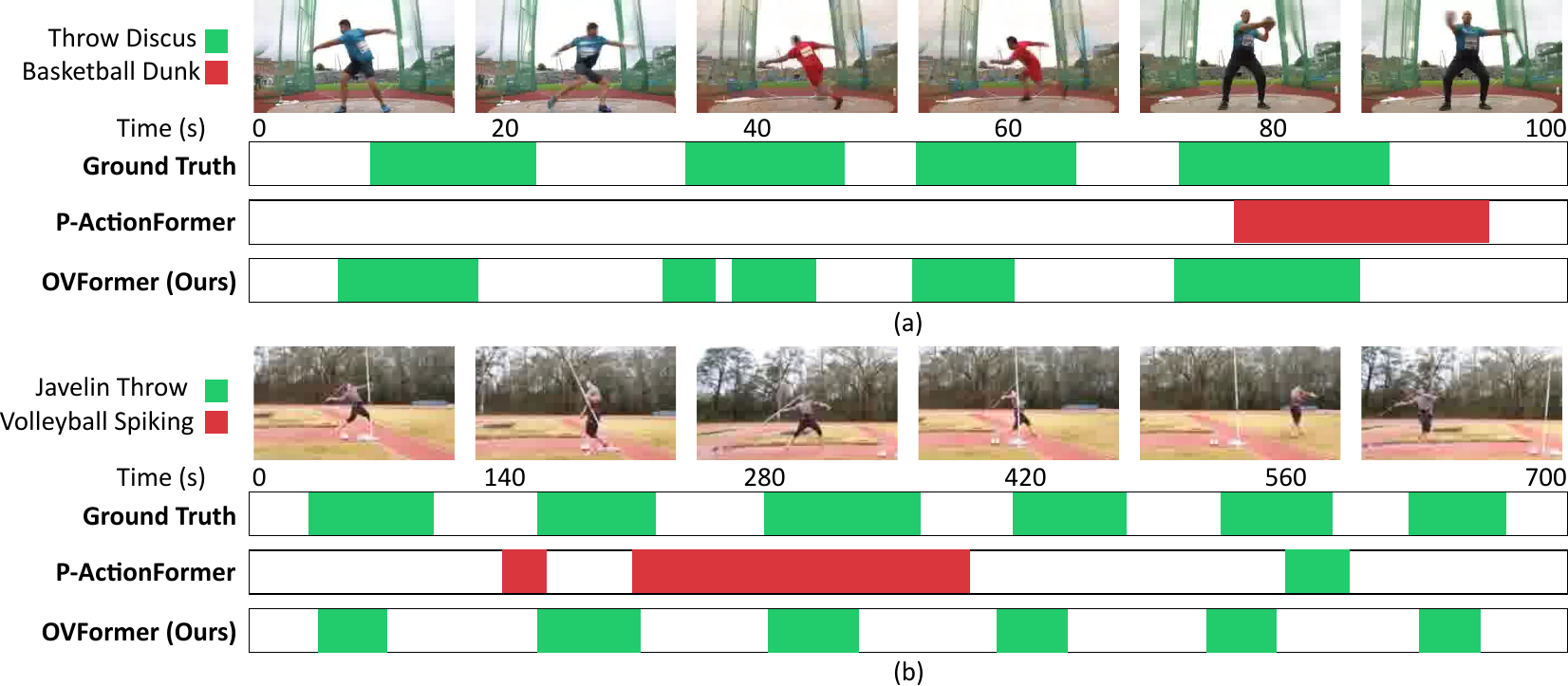}
    \vspace{0.1cm}
    \caption{\small \textbf{OVTAL comparison between \ovformer and P-\aformer on the test set for \thumos with a 50-50 split on base and novel action categories.} The top row shows the ground truth action boundaries, the middle row shows the baseline method P-\aformer's performance, and the bottom row shows the performance of our proposed method \ovformer. In (a), P-\aformer struggles to differentiate between the novel action category \texttt{Throw Discus} and the base action category \texttt{Basketball Dunk}. Similarly, in (b), P-\aformer confuses the novel action category \texttt{Javelin Throw} with the base action category \texttt{Volleyball Spiking}. These errors occur due to the visual similarities between the action categories.
    In contrast, our proposed method is able to correctly localize the action boundaries. See \autoref{sec:suppl_gentext_th} for more details.}

    \label{fig:thumos_gzsl}
\end{figure*}

\begin{figure*}[t]
    \centering
    \includegraphics[width=1.0\linewidth]{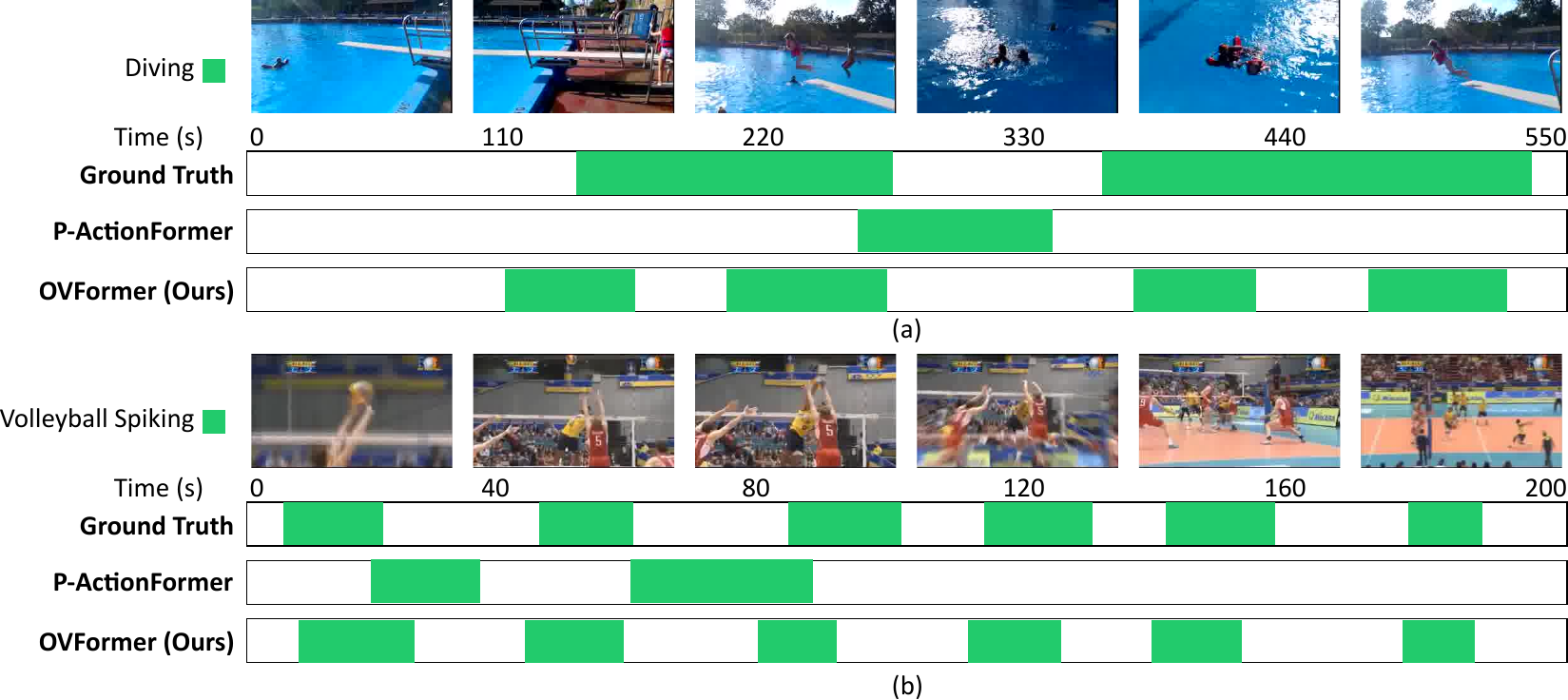}
    \vspace{0.1cm}
    \caption{\small \textbf{OVTAL comparison between \ovformer and P-\aformer on the test set for \thumos with a 50-50 split on novel action categories.} The top row shows the ground truth action boundaries, the middle row shows the baseline method P-\aformer performance, and the bottom row shows the performance for our proposed method \ovformer. We can see that P-\aformer misses the action boundaries for the ground-truth classes \texttt{Diving} in (a) and \texttt{Volleyball Spiking} in (b) whereas our proposed method is able to localize the action boundaries correctly. See \autoref{sec:suppl_gentext_th} for more details.}
    \label{fig:thumos_zsl}
\end{figure*}

\begin{figure*}[t] 
    \centering
    \includegraphics[width=1.0\linewidth]{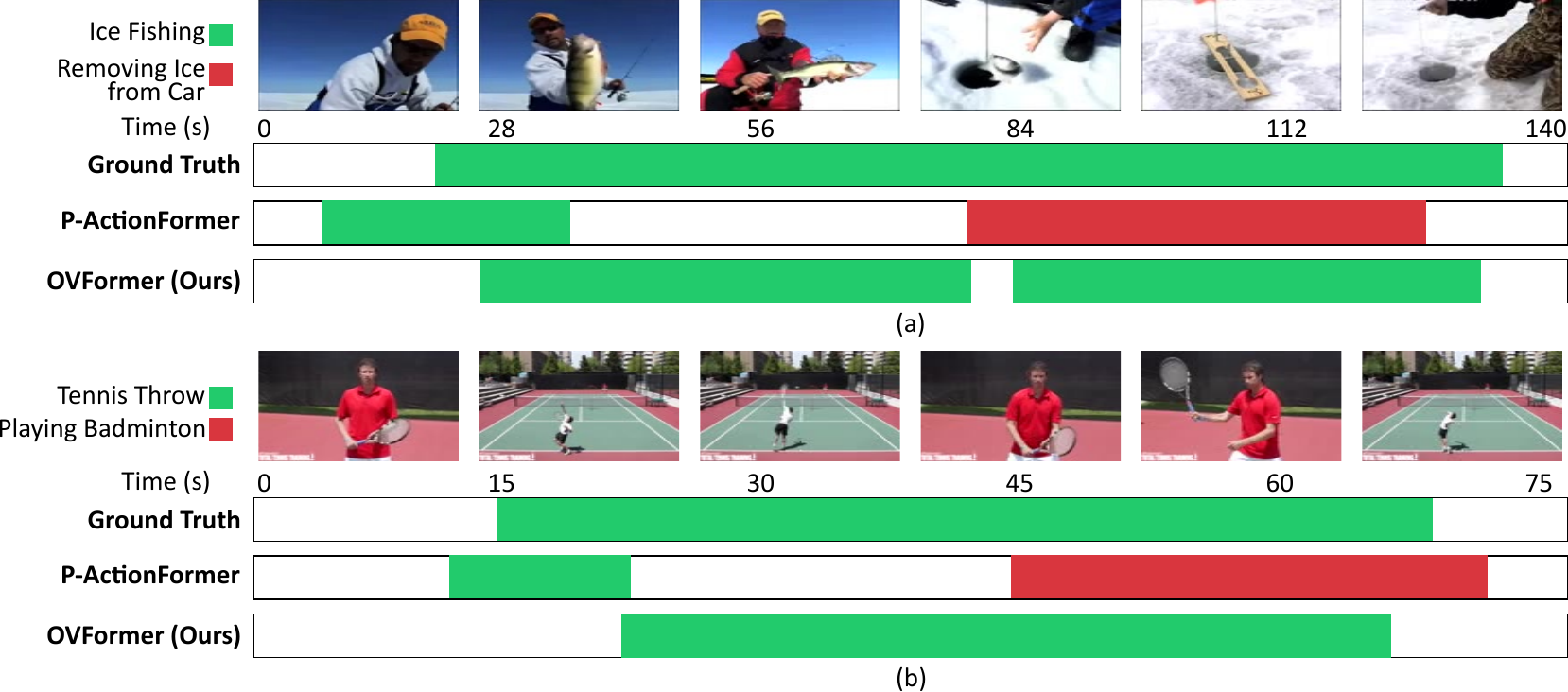}
    \vspace{0.1cm}
    \caption{\small \textbf{OVTAL comparison between \ovformer and P-\aformer on the test set for \anet with a 50-50 split on base and novel action categories.} The top row shows the ground truth action boundaries, the middle row shows the baseline method P-\aformer performance, and the bottom row shows the performance for our proposed method \ovformer. In (a), P-\aformer struggles to differentiate between the novel action category \texttt{Removing Ice from Car} and the base action category \texttt{Ice Fishing}. Similarly, in (b), P-\aformer confuses the novel action category \texttt{Tennis Throw} with the base action category \texttt{Playing Badminton}. These errors occur due to the visual similarities between the action categories. Our proposed method is able to localize the action boundaries correctly. See \autoref{sec:suppl_gentext_anet} for more details.}
    \label{fig:anet_gzsl}
\end{figure*}

\begin{figure*}[t]
    \centering
    \includegraphics[width=1.0\linewidth]{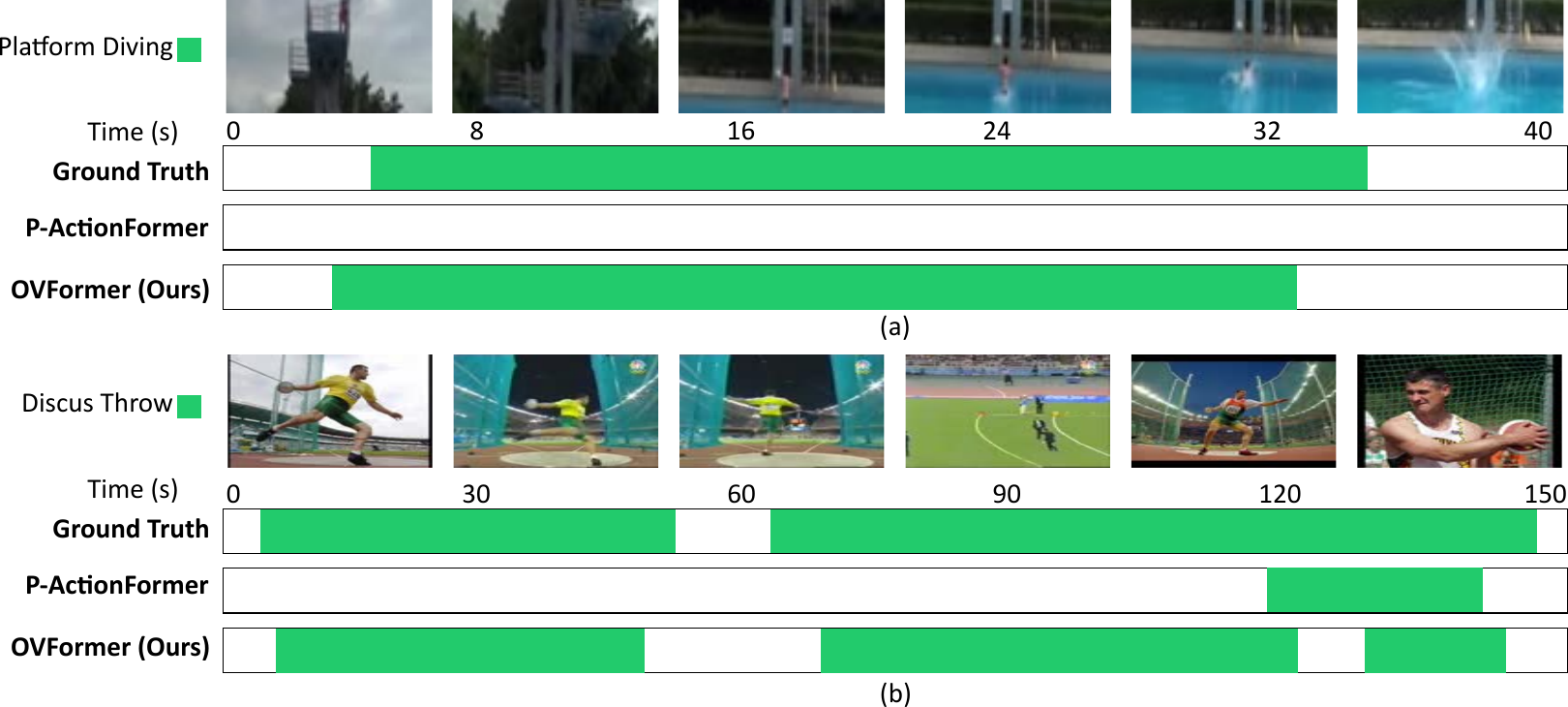}
    \vspace{0.1cm}
    \caption{\small \textbf{OVTAL comparison between \ovformer and P-\aformer on the test set for \anet with a 50-50 split on novel action categories.} The top row shows the ground truth action boundaries, the middle row shows the baseline method P-\aformer performance, and the bottom row shows the performance for our proposed method \ovformer. We can see that P-\aformer misses the action boundaries for the ground-truth classes \texttt{Platform Diving} in (a) and \texttt{Discus Throw} in (b) whereas our proposed method is able to localize the action boundaries correctly. See \autoref{sec:suppl_gentext_anet} for more details. }
    \label{fig:anet_zsl}
\end{figure*}

In this section, we show additional qualitative results comparing the performance of \ovformer to the baseline method P-\aformer on the \thumos and \anet datasets. We show results for both novel action categories (\autoref{fig:thumos_zsl} and \autoref{fig:anet_zsl}) and base and novel action categories (\autoref{fig:thumos_gzsl} and \autoref{fig:anet_gzsl}). In each figure, the top row displays the ground truth action boundaries, the middle row shows the predictions from P-\aformer, and the bottom row presents the predictions from \ovformer. We observe that \ovformer improves localization performance for novel action categories compared to P-\aformer. Specifically, in \autoref{fig:thumos_gzsl}(a), which shows results on base and novel action categories from \thumos, P-\aformer confuses \texttt{Throw Discus} (novel class) and \texttt{Basketball Dunk} (base class) actions when the body movements hold a very strong similarity. However, \ovformer can correctly separate these action categories, showing the significance of the multimodal guided features that capture rich scene information and semantic context related to the actions. Furthermore, in \autoref{fig:thumos_gzsl}(b), also on \thumos, P-\aformer confuses \texttt{Javelin Throw} (base class) and \texttt{Volleyball Spiking} (novel class) actions, while \ovformer can correctly distinguish between them. In \autoref{fig:thumos_zsl}, which shows results on novel action categories from \thumos, P-\aformer misses the action boundaries for the ground-truth classes \texttt{Diving} (\autoref{fig:thumos_zsl}(a)) and \texttt{Volleyball Spiking} (\autoref{fig:thumos_zsl}(b)), whereas \ovformer is able to correctly localize the action boundaries.

On the \anet dataset, \autoref{fig:anet_gzsl}  shows the localization comparison between \ovformer and P-\aformer on base and novel action categories. In \autoref{fig:anet_gzsl}(a), P-\aformer gets confused between visually similar action categories, such as \texttt{Ice Fishing} (base class) and \texttt{Removing Ice from Car} (novel class), leading to inaccurate localization of the action boundaries when the action category holds visual similarity with other action categories. 
Similarly, in \autoref{fig:anet_gzsl}(b), P-\aformer confuses \texttt{Tennis Throw} (novel class) and \texttt{Playing Badminton} (base class), while \ovformer can correctly distinguish between them.
In \autoref{fig:anet_zsl}, which shows results on novel action categories from \anet, P-\aformer misses the action boundaries for the ground-truth classes \texttt{Platform Diving} (\autoref{fig:anet_zsl}(a)) and \texttt{Discus Throw} (\autoref{fig:anet_zsl}(b)), whereas \ovformer is able to correctly localize the action boundaries.
All these qualitative examples demonstrate \ovformer's strong open-vocabulary capability, as it leverages multimodal representations to effectively recognize and localize novel action categories that were unseen during training. This is in contrast to P-\aformer, which struggles to distinguish between visually similar actions, especially for novel categories.

\begin{figure*}[t]
    \centering
    \includegraphics[width=1.0\linewidth]{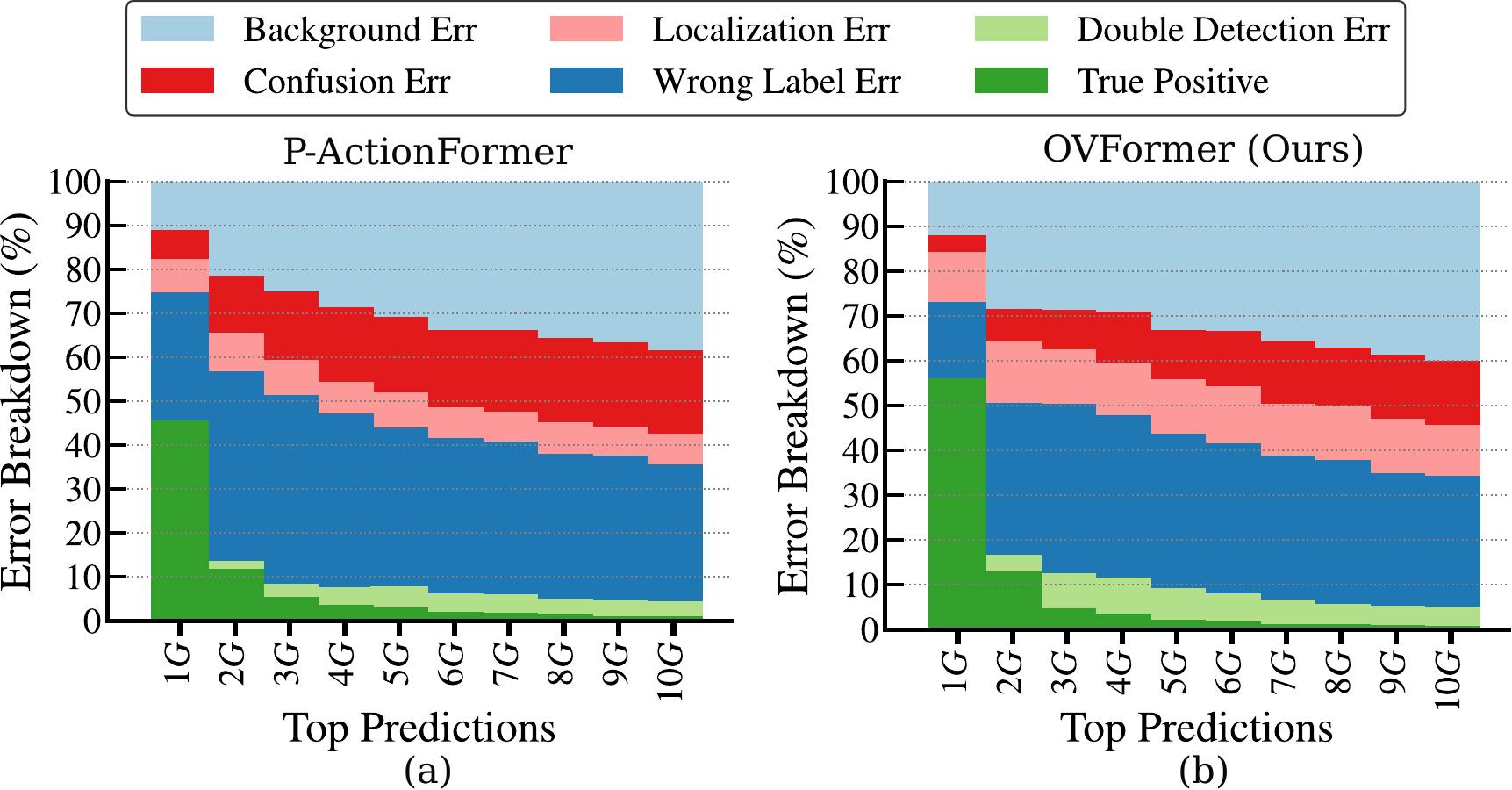}
    \vspace{0.1cm}
    \caption{False positive (FP) profiling on \thumos on 50-50 split using the approach from \cite{alwassel2018diagnosing}. The figure shows the FP error breakdown for the top 10 ground truth predictions per action category. On the left (a), we have the baseline method P-\aformer, and on the right (b), we present our proposed method \ovformer. We observe a significant improvement in true positives for our proposed method and a substantial decrease in confusion errors compared to the baseline method.}
    \label{fig:thumos_gzsl_5050_fp}
\end{figure*}

In \autoref{fig:thumos_gzsl_5050_fp}, we perform a false positive (FP) analysis at tIOU=0.5 for \thumos for 50-50 split on base and novel action categories. For clarity, we choose to show the results on one of the splits from the 10 random splits. We compare the baseline method P-\aformer (\autoref{fig:thumos_gzsl_5050_fp}(a)) and \ovformer (\autoref{fig:thumos_gzsl_5050_fp}(b)). We can see a significant improvement in true positive prediction which clearly shows the significance of Stage \RNum{1} training on a larger vocabulary dataset and multimodal guided features for OVTAL. For more detailed explanations regarding the FP analysis chart and error categorization, we refer the readers to the work \cite{alwassel2018diagnosing}, which introduced this diagnostic tool for evaluating temporal action localization models.


\clearpage

\section{Generated Class Description Examples: \thumos \label{sec:suppl_gentext_th}}

In this section, we show 10 rich text descriptions generated using the \texttt{gpt-3.5-turbo-instruct} model from OpenAI for five classes from \thumos. All text descriptions will be released publicly along with the code.

\subsection{Generated Description for `BaseballPitch':}
\begin{enumerate}
        \footnotesize
        \setlength{\itemsep}{0pt}%
        \setlength{\parskip}{0pt}%
        \item \texttt{You can recognize a video of a person performing the BaseballPitch action by looking for certain key actions such as a raised arm, a windup involving a back and forth motion of the arms and a follow-through, a powerful transfer of weight.
        } \item \texttt{ A video of a person performing a BaseballPitch action can be recognized by the player throwing the baseball with their arm, with their body facing forward and their arm in a slightly bent position, and then releasing the ball with a snapping motion of.
        } \item \texttt{ The most reliable way to recognize a video of a person performing a BaseballPitch action is by looking for certain visual cues.
        } \item \texttt{ These cues include the pitcher raising their leg in a kicking motion, a forward-leaning torso, arms bent at a 90.
        } \item \texttt{ A video of a person performing a BaseballPitch action can be recognized by looking for the following clues: the person holding the ball in an overhand grip, bringing the arm back with the elbow raised, cocking the wrist, and then.
        } \item \texttt{ A video of a person performing a Baseball Pitch action can be recognized by looking for certain movements in the video.
        } \item \texttt{ Key features of a Baseball Pitch include the pitcher winding up by swinging backwards with their arm, bringing their body straight, and then bringing.
        } \item \texttt{ A video of someone performing a Baseball Pitch action can be identified by looking for a sequence of distinct motions.
        } \item \texttt{ These motions should include the windup, transitioning to the leg kick, driving their arm towards the plate, and releasing the ball.
        } \item \texttt{ You can recognize a video of a person performing a Baseball Pitch action by looking for features such as arm movement in the windup position, a smooth overhand delivery, and the follow\-through of the pitch.
    }
    \end{enumerate}

\subsection{Generated Description for `CliffDiving':}

\begin{enumerate}
        \footnotesize
        \setlength{\itemsep}{0pt}%
        \setlength{\parskip}{0pt}%
            \item \texttt{ One way to recognize a video of a person performing CliffDiving action is by looking for the following visual cues: a high elevation from the ground, a person diving from the cliff, and either a pool, lake, or ocean nicely situated below.
            } \item \texttt{ Cliff diving can be easily identified by looking for a person performing high jumps and dives off a high cliff into the water below.
            } \item \texttt{ The cliff diving locations will generally have a steep drop off which is why it is considered a high-risk sport.
            } \item \texttt{ A video of a person performing a CliffDiving action can be recognized by looking for key traits of cliff diving, such as jumping off a cliff, performing a flip or spin, and entering the water feet first.
            } \item \texttt{ Cliff diving is an extreme sport that involves diving off a cliff or other high structure into water.
            } \item \texttt{ To recognize a video of a person performing a cliff diving action, look for visuals of a person leaping off a high structure into water and flashing.
            } \item \texttt{ You can recognize a video of someone performing a cliff diving action by looking for clues such as a high cliff or outcropping of rock, a person in swimming gear or a wet suit, and the person leaping into the water from the cliff.
            } \item \texttt{ A video of someone performing a CliffDiving action would typically involve a person diving off of a tall cliff or precipice into the water below.
            } \item \texttt{ In the video, you may see the person taking a running start, executing a somersault.
            } \item \texttt{ A video of a person performing a CliffDiving action can be recognized by looking for visuals of an individual jumping and/or diving off a high cliff into a body of water.
            }
    \end{enumerate}

\subsection{Generated Description for `FrisbeeCatch':}

\begin{enumerate}
        \footnotesize
        \setlength{\itemsep}{0pt}%
        \setlength{\parskip}{0pt}%
        \item \texttt{ You can recognize a video of a person performing FrisbeeCatch action by looking for the motions of throwing and catching a Frisbee in the video.
        } \item \texttt{ You should also look for visual cues such as the Frisbee itself and any.
        } \item \texttt{ To recognize a video of a person performing the FrisbeeCatch action, look for the following visual cues: the individual throwing the Frisbee, the Frisbee in the air, the person catching the Frisbee, and.
        } \item \texttt{ A person performing the FrisbeeCatch action can be recognized by their stance – a low athletic position ready to catch the fly-by disc, and by the way they’re moving – arms outstretched and eyes tracking the fr.
        } \item \texttt{ A video of someone performing the Frisbee Catch action can be recognized by seeing them throw a frisbee in the air, and then quickly running to catch it before it hits the ground.
        } \item \texttt{ There should also be an obvious throwing and catching.
        } \item \texttt{ You can recognize a video of someone performing FrisbeeCatch by looking for one or more persons throwing and catching a Frisbee.
        } \item \texttt{ It should be clear that the persons are attempting to catch the Frisbee while it is in the.
        } \item \texttt{ You can recognize a video of someone performing the FrisbeeCatch action by looking for images of someone throwing a Frisbee and watching to see if they catch it in their hands.
        } \item \texttt{ Additionally, the video should include the person running, jumping and stretching to catch the Frisbee.
        }
    \end{enumerate}

\subsection{Generated Description for `JavelinThrow':}

\begin{enumerate}
        \footnotesize
        \setlength{\itemsep}{0pt}%
        \setlength{\parskip}{0pt}%
        \item \texttt{ In a video of someone performing the JavelinThrow action, you should look for a person throwing a javelin with good technique and form as well as the javelin leaving their hands and flying through the air.
        } \item \texttt{ A video of a person performing a JavelinThrow action can be recognized by observing the person’s technique as they grip the javelin tightly in their hand, run towards the throwing line, and hurl the javelin.
        } \item \texttt{ You can recognize a video of a person performing the JavelinThrow action by looking for visual clues.
        } \item \texttt{ The presence of a javelin in the video
        } \item \texttt{ A person gripping the javelin, winding.
        } \item \texttt{ You can recognize a video of a person performing a JavelinThrow action if you observe the person holding a javelin in their dominant hand and throwing it with their arm up in an arching motion.
        } \item \texttt{ You may also see them run.
        } \item \texttt{ You can recognize a video of person performing JavelinThrow action by looking for certain key elements.
        } \item \texttt{ These elements include a person gripping the javelin, running down the field/track, throwing the javelin and watching it soar.
        } \item \texttt{ The person will have a javelin in their hand.
        }
    \end{enumerate}

\subsection{Generated Description for `Billiards':}

\begin{enumerate}
        \footnotesize
        \setlength{\itemsep}{0pt}%
        \setlength{\parskip}{0pt}%
        \item \texttt{ You can recognize a video of a person performing billiards action by looking for the visual cues of a billiards table, the holding and playing of the billiards cue by the person, and the striking of the billiard balls.
        } \item \texttt{ To recognize a video of a person performing billiards action, you can look for cues such as a pool table or billiards equipment, a person holding a billiards cue, shots of the ball impacting other balls or the cushion.
        } \item \texttt{ You can recognize a video of a person performing billiards action by looking for a number of visuals.
        } \item \texttt{ These visuals could include a person gripping a pool cue, a pool table, pool balls, and objects being struck by the cue ball.
        } \item \texttt{ A video of someone performing a billiards action can be identified by seeing them take shots with a standard billiards cue stick, striking balls on the billiards table and positioning them with the use of their cue stick, as well.
        } \item \texttt{ To recognize a video of someone performing billiards action, you should look for the following elements: a billiards table with two pockets at either end, cues or pool sticks and a complete set of billiard balls.
        } \item \texttt{ You can recognize a video of a person performing billiards action by looking for cues such as the cue ball, pool stick, and billiard table, as well as watching for specific movements associated with billiards.
        } \item \texttt{ To recognize a video of a person performing a billiards action, you should look for recognizable cues such as the person grabbing a pool cue, the sound of a ball being hit, and the movement of balls on the table.
        } \item \texttt{ You can recognize a video of a person performing Billiards action by looking for cues such as the person holding a pool cue, a pool table with the balls arranged in a rack, and the sound of the balls being struck together during the.
        } \item \texttt{ To recognize a video of person performing billiards action, look for cues such as the billiard table, billiard balls, cues, and the various motions of the person playing the game.
        }
    \end{enumerate}

\section{Generated Class Description Examples: \anet \label{sec:suppl_gentext_anet}}

In this section, we show 10 rich text descriptions generated using the \texttt{gpt-3.5-turbo-instruct} model from OpenAI for five classes from \anet. All text descriptions will be released publicly along with the code.

\subsection{Generated Description for `Applying sunscreen':}

\begin{enumerate}
        \footnotesize
        \setlength{\itemsep}{0pt}%
        \setlength{\parskip}{0pt}%
        \item \texttt{ You can recognize a video of a person performing the action of applying sunscreen by watching them slather the sunscreen on their skin, rubbing it in until their skin is covered, and seeing them put the sunscreen away when they are finished.
        } \item \texttt{ You can recognize a video of person performing the action of applying sunscreen by looking for the typical signs of the action.
        } \item \texttt{ You can recognize a video of a person performing the action of applying sunscreen if the person is seen taking out a topical sunscreen product from its container, then applying the product to their skin, ensuring that all exposed skin areas are covered.
        } \item \texttt{ If you are looking for a video of someone performing the action of applying sunscreen, you may search for terms such as "applying sunscreen video", "sunscreen application", or "sunscreen application tutorial".
        } \item \texttt{ You can recognize a video of a person performing the action of applying sunscreen by looking for visual cues such as the person applying a white creamy sunscreen product to their face, ears, arms, legs, etc.
        } \item \texttt{ Visual cues you may look out for in a video of a person applying sunscreen may include seeing someone's hands applying lotion or cream onto their exposed skin, rubbing the lotion into the skin, and/or seeing the person use a sun.
        } \item \texttt{ You can recognize a video of a person applying sunscreen action by looking for someone taking out a bottle of sunscreen from a bag and then applying it to exposed skin.
        } \item \texttt{ You can recognize a video of a person performing the action of applying sunscreen by looking for certain items used when applying sunscreen.
        } \item \texttt{ The video could show the person taking sunscreen in the palm of their hand and applying it on their skin.
        } \item \texttt{ You can recognize a video of a person performing the action of applying sunscreen by looking for visual cues such as a person of any age, gender, or ethnicity.
        }
    \end{enumerate}

\subsection{Generated Description for `Braiding hair':}

\begin{enumerate}
        \footnotesize
        \setlength{\itemsep}{0pt}%
        \setlength{\parskip}{0pt}%
        \item \texttt{ You can recognize a video of someone performing Braiding hair by looking for someone with a comb in their hand who is separating the hair into sections, twisting the sections of hair around each other and securing each section with a hair tie or clip.
        } \item \texttt{ You can recognize a video of someone performing the Braiding Hair action by looking for distinct movements such as: sectioning the hair into 3 or more sections, crossing the outer sections over the inner section, looping the strands around each other,.
        } \item \texttt{ You can recognize a video of person performing braiding hair action by looking for someone holding several strands of hair, parting it into sections, and weaving them into a tight plait or braid.
        } \item \texttt{ You can recognize a video of someone performing a braiding hair action by looking for visual cues, such as images of someone with their hands braiding another person's hair and/or visible motion of someone's hands doing a braid.
        } \item \texttt{ Look for a video that shows a person with their hands weaving together strands of hair.
        } \item \texttt{ You can recognize a video of a person performing the Braiding Hair action by looking for someone who is using their hands to weave and braid hair strands together and forming patterns.
        } \item \texttt{ You can recognize a video of a person performing Braiding hair action by looking for a person with their hands moving back and forth as if they are weaving together sections of hair.
        } \item \texttt{ You can recognize a video of a person performing the braiding hair action by looking for someone separation sections of the hair with their hands and weaving them together over and over to create a woven pattern.
        } \item \texttt{ You can recognize a video of someone performing braiding hair by looking for visual indications of the person or people in the video performing the action of braiding hair.
        } \item \texttt{ You can recognize a video of a person performing a Braiding hair action by looking for specific visuals such as a person with their hair parted in the middle, with three strands of hair taken and twisted together in a specific pattern.
        }
    \end{enumerate}

\subsection{Generated Description for `Drinking coffee':}

\begin{enumerate}
        \footnotesize
        \setlength{\itemsep}{0pt}%
        \setlength{\parskip}{0pt}%
        \item \texttt{ The person will typically be seen stirring or mixing their coffee, picking up the mug and bringing it to their mouth, and drinking from the mug.
        } \item \texttt{ You can recognize a video of someone drinking coffee by looking for visual cues such as someone picking up a cup, pushing a lid off of a cup, pouring a liquid into a cup, or putting a spoonful of sugar into a cup.
        } \item \texttt{ You can recognize a video of someone drinking coffee by looking for certain visuals and sounds.
        } \item \texttt{ You can look for video footage of the person holding a coffee cup, drinking from the cup, or stirring the coffee with a spoon.
        } \item \texttt{ You can recognize a video of someone performing the action of drinking coffee by looking for familiar motions, like lifting a cup to their lips, and the characteristic sound of a person savoring a sip of hot drink.
        } \item \texttt{ A video of a person performing the Drinking Coffee action can be recognized by visual cues, such as the person picking up a mug, bringing the mug to their lips, and then taking a sip of coffee.
        } \item \texttt{ You can recognize a video of a person performing the drinking coffee action by looking for visual cues such as the person holding a mug, steam rising from a cup, and/or the person taking a sip of the coffee.
        } \item \texttt{ You could look for video footage of someone taking a sip of coffee, preparing coffee, or pouring coffee into a cup.
        } \item \texttt{ You can recognize a video of a person performing the Drinking Coffee action by looking for the action of a person picking up a cup of coffee and putting it to their mouth.
        } \item \texttt{ You can recognize a video of a person performing the Drinking coffee action by looking for visuals such as a person holding a mug of coffee, making the drinking motion with their hand, or looking into a cup of coffee.
        }
    \end{enumerate}

\subsection{Generated Description for `Skiing':}

\begin{enumerate}
        \footnotesize
        \setlength{\itemsep}{0pt}%
        \setlength{\parskip}{0pt}%
        \item \texttt{ A video of someone performing a skiing action can be recognized by observing how the person moves their body and skis down a slope.
        } \item \texttt{ You can recognize a video of a person performing a skiing action by looking for recognizable ski clothing, skis, ski poles and other ski equipment, and by watching for the person to make recognizable skiing motions, such as gliding down a hill.
        } \item \texttt{ You can recognize a video of someone performing a skiing action by looking for the following elements: the person wearing ski apparel, the skiing equipment and the environment (snow-covered slopes, ski-lifts, and other skiers).
        } \item \texttt{ One way to recognize a video of someone performing the skiing action is to look for telltale signs such as the person wearing alpine skiing equipment, such as ski boots, skis, poles, and a helmet.
        } \item \texttt{ Look for someone skiing down a hill with skis, poles, and ski boots.
        } \item \texttt{ You may recognize a video of someone skiing by looking for recognizable skiing positions and movements, such as edging, carving, and making turns.
        } \item \texttt{ One way to recognize a video of a person performing the skiing action is to look for clues such as snow, skis, ski poles, and the crouched position that a skier assumes when skiing.
        } \item \texttt{ You can recognize a video of person performing skiing action by looking for visual elements that include a person skiing down a slope or off a jump and make turns, wearing ski equipment like boots, bindings, and skis.
        } \item \texttt{ You can recognize a video of someone performing skiing action by looking for specific visual cues.
        } \item \texttt{ You can recognize a video of someone performing skiing by looking for recognizable skiing movements such as a two-footed gliding motion, making turns in the snow, or controlling speed by using pole plants.
        }
    \end{enumerate}

\subsection{Generated Description for `Making a sandwich':}

\begin{enumerate}
        \footnotesize
        \setlength{\itemsep}{0pt}%
        \setlength{\parskip}{0pt}%
        \item \texttt{ You can recognize a video of person performing the action of Making a sandwich by looking for visual clues such as seeing a person assembling bread, meat, cheese, and other ingredients; slicing these ingredients; and arranging them on a plate.
        } \item \texttt{ You can recognize a video of a person performing the action of making a sandwich by observing the physical movement of the person putting ingredients between two slices of bread, such as meat, cheese, and condiments, and then finishing off the process by.
        } \item \texttt{ You can recognize a video of someone making a sandwich by looking for footage of them putting bread, meat, and vegetables onto a plate and combining them into a sandwich.
        } \item \texttt{ You can recognize a video of a person performing the action of making a sandwich by observing the person going through the steps of constructing the sandwich, such as spreading the condiments, arranging the ingredients, and slicing the sandwich in half.
        } \item \texttt{ You can recognize a video of someone performing the action of making a sandwich by looking for visual cues such as a person cutting, spreading, and arranging various ingredients on bread or an alternative base.
        } \item \texttt{ You can recognize a video of a person making a sandwich by looking for several key components.
        } \item \texttt{ You can recognize a video of a person making a sandwich by observing the visual of the person assembling the sandwich, such as spreading butter, putting slices of meat and cheese, adding condiments and vegetables, then cutting it in half.
        } \item \texttt{ You can recognize a video of someone making a sandwich action by looking for someone with bread, fillings, and any other necessary items such as knives, cutting boards, etc.
        } \item \texttt{ You can recognize a video of a person performing the action of making a sandwich by looking for visuals such as: someone assembling two pieces of bread, adding condiments such as meat, cheese and/or vegetables, and putting condiments like mayo.
        } \item \texttt{ To recognize a video of a person performing the action of making a sandwich, you can look for visuals of the person gathering the ingredients for a sandwich, assembling the sandwich together, and then cutting the sandwich into slices.
        }
    \end{enumerate}


\end{document}